# Joint Estimation of Multiple Graphical Models from High Dimensional Time Series

Huitong Qiu*, Fang Han †, Han Liu ‡, and Brian Caffo §

October 8, 2014

## Abstract

In this manuscript we consider the problem of jointly estimating multiple graphical models in high dimensions. We assume that the data are collected from $n$ subjects, each of which consists of $T$ possibly dependent observations. The graphical models of subjects vary, but are assumed to change smoothly corresponding to a measure of closeness between subjects. We propose a kernel based method for jointly estimating all graphical models. Theoretically, under a double asymptotic framework, where both $(T, n)$ and the dimension $d$ can increase, we provide the explicit rate of convergence in parameter estimation. It characterizes the strength one can borrow across different individuals and impact of data dependence on parameter estimation. Empirically, experiments on both synthetic and real resting state functional magnetic resonance imaging (rs-fMRI) data illustrate the effectiveness of the proposed method.

*Department of Biostatistics, Johns Hopkins University, Baltimore, MD 21205, USA; e-mail: hqiu@jhsph.edu

†Department of Biostatistics, Johns Hopkins University, Baltimore, MD 21205, USA; e-mail: fhan@jhsph.edu

‡Department of Operations Research and Financial Engineering, Princeton University, Princeton, NJ 08544, USA; e-mail: hanliu@princeton.edu

§Department of Biostatistics, Johns Hopkins University, Baltimore, MD 21205, USA; e-mail: bcaffo@gmail.com

# 1 Introduction

Undirected graphical models encoding the conditional independence structure among the variables in a random vector have been heavily exploited in multivariate data analysis (Lauritzen, 1996). In particular, when $\boldsymbol{X} \sim N_d(\boldsymbol{0}, \boldsymbol{\Sigma})$ is a $d$ dimensional Gaussian vector, estimating such graphical models is equivalent to estimating the nonzero entries in the inverse covariance matrix $\boldsymbol{\Theta} := \boldsymbol{\Sigma}^{-1}$ (Dempster, 1972). The undirected graphical model encoding the conditional independence structure for the Gaussian distribution is sometimes called a Gaussian graphical model.

There has been much work on estimating a single Gaussian graphical model, $\mathbf{G}$, based on $n$ independent observations. In low dimensional settings where the dimension, $d$, is fixed, Drton and Perlman (2007) and Drton and Perlman (2008) proposed to estimate $\mathbf{G}$ using multiple testing procedures. In settings where the dimension, $d$, is much larger than the sample size, $n$, Meinshausen and Bühlmann (2006) proposed to estimate $\mathbf{G}$ by solving a collection of regression problems via the lasso. Yuan and Lin (2007), Banerjee et al. (2008), Friedman et al. (2008), Rothman et al. (2008), and Liu and Luo (2012) proposed to directly estimate $\boldsymbol{\Theta}$ using the $\ell_1$ penalty (detailed definition provided later). More recently, Yuan (2010) and Cai et al. (2011) proposed to estimate $\boldsymbol{\Theta}$ via linear programming. The above mentioned estimators are all consistent with regard to both parameter estimation and model selection, even when $d$ is nearly exponentially larger than $n$.

This body of work is focused on estimating a single graph based on independent realizations of a common random vector. However, in many applications this simple model does not hold. For example, the data can be collected from multiple individuals that share the same set of variables, but differ with regard to the structures among variables. This situation is frequently encountered in the area of brain connectivity network estimation (Friston, 2011). Here brain connectivity networks corresponding to different subjects vary, but are expected to be more similar if the corresponding subjects share many common demographic, health or other covariate features. Under this setting, estimating the graphical models separately for each subject ignores the similarity between the adjacent graphical models. In contrast, estimating one population graphical model based on the data of all subjects ignores the differences between graphs and may lead to inconsistent estimates.

There has been a line of research in jointly estimating multiple Gaussian graphical models for independent data. On one hand, Guo et al. (2011) and Danaher et al. (2014) proposed methods via introducing new penalty terms, which encourage the sparsity of both the parameters in each subject and the differences between parameters in different subjects. On the other hand, Song et al. (2009a), Song et al. (2009b), Kolar and Xing (2009), Zhou et al. (2010), and Kolar et al. (2010) focused on independent data with time-varying networks. They proposed efficient algorithms for estimating and predicting the networks along the time line.

In this paper, we propose a new method for jointly estimating and predicting networks corresponding to multiple subjects. The method is based on a different model compared to the ones listed above. The motivation of this model arises from resting state functional magnetic resonance imaging (rs-fMRI) data, where there exist many natural orderings corresponding to measures of health status, demographics, and many other subject-specific covariates. Moreover, the observations of each subject are multiple brain scans with temporal dependence. Accordingly, different from the methods in estimating time varying networks, we need to handle the data where each subject has $T$, instead of one, observations. Different from the methods in Guo et al. (2011) and Danaher et al. (2014), it is assumed that there exists a natural ordering for the subjects, and the parameters of interest vary smoothly corresponding to this ordering. Moreover, we allow the observations to be dependent via a temporal dependence structure. Such a setting has not been studied in high dimensions until very recently (Loh and Wainwright, 2012; Han and Liu, 2013; Wang et al., 2013).

We exploit a similar kernel based approach as in Zhou et al. (2010). It is shown that our method can efficiently estimate and predict multiple networks while allowing the data to be dependent. Theoretically, under a double asymptotic framework, where both $d$ and $(T, n)$ may increase, we provide an explicit rate of convergence in parameter estimation. It sharply characterizes the strength one can borrow across different subjects and the impact of data dependence on the convergence rate. Empirically, we illustrate the effectiveness of the proposed method on both synthetic and real rs-fMRI data. In detail, we conduct comparisons of the proposed approach with several existing methods under three synthetic patterns of evolving graphs. In addition, we study the large scale ADHD-200 dataset to investigate the development of brain connectivity networks over age, as well as the effect of kernel bandwidth on estimation, where scientifically interesting results are unveiled.

We note that the proposed multiple time series model has analogous prototypes in spatial-temporal analysis. This line of work is focused on multiple times series indexed by a spatial variable. A common strategy models the spatial-temporal observations by a joint Gaussian process, and imposes a specific structure on the spatial-temporal covariance function (Jones and Zhang, 1997; Cressie and Huang, 1999). Another common strategy decomposes the temporal series into a latent spatial-temporal structure and a residual noise. Examples of the latent spatial-temporal structure include temporal autoregressive processes (Høst et al., 1995; Sølna and Switzer, 1996; Antunes and Rao, 2006; Rao, 2008) and mean processes (Storvik et al., 2002; Gelfand et al., 2003; Banerjee et al., 2004, 2008; Nobre et al., 2011). The residual noise is commonly modeled by a parametric process such as a Gaussian process. The aforementioned literature is restricted in three aspects. First, they only consider univariate or low dimensional multivariate spatial-temporal series. Secondly, they restrict the covariance structure of the observations to a specific form. Thirdly, none of this literature addresses the problem of estimating the conditional independence structure of the time series. In comparison, we consider estimating the conditional independence graph under high dimensional times series. Moreover, our model involves no assumption on the structure of the covariance matrix.

We organize the rest of the paper as follows. In Section 2, the problem setup is introduced and the proposed method is given. In Section 3, the main theoretical results are provided. In Section 4, the method is applied to both synthetic and rs-fMRI data to illustrate its empirical usefulness. A discussion is provided in the last section, while detailed technical proofs are put in the appendix.

## 2 The Model and Method

Let $\mathbf{M} = (M_{jk}) \in \mathbb{R}^{d \times d}$ and $\boldsymbol{v} = (v_1, ..., v_d)^T \in \mathbb{R}^d$. We denote $\boldsymbol{v}_I$ to be the subvector of $\boldsymbol{v}$ whose entries are indexed by a set $I \subset \{1, \ldots, d\}$. We denote $\mathbf{M}_{I,J}$ to be the submatrix of $\mathbf{M}$ whose rows are indexed by $I$ and columns are indexed by $J$. Let $\mathbf{M}_{I,*}$ be the submatrix of $\mathbf{M}$ whose rows are indexed by $I$, and $\mathbf{M}_{*,J}$ be the submatrix of $\mathbf{M}$ whose columns are indexed by $J$. For $0 < q < \infty$, define the $\ell_0$, $\ell_q$, and $\ell_\infty$ vector norms as

$$\|\boldsymbol{v}\|_0 = \sum_{j=1}^d I(v_j \neq 0), \quad \|\boldsymbol{v}\|_q := \left(\sum_{j=1}^d |v_j|^q\right)^{1/q}, \quad \text{and} \quad \|\boldsymbol{v}\|_\infty = \max_{1 \leq j \leq d} |v_j|,$$

where $I(\cdot)$ is the indicator function. For a matrix $\mathbf{M}$, denote the matrix $\ell_q$, $\ell_{\max}$, and Frobenius norms to be

$$\|\mathbf{M}\|_q = \max_{\|\boldsymbol{v}\|_q=1} \|\mathbf{M}\boldsymbol{v}\|_q, \quad \|\mathbf{M}\|_{\max} = \max_{jk} |M_{jk}|, \quad \text{and} \quad \|\mathbf{M}\|_F = \left(\sum_{j,k} |M_{jk}|^2\right)^{1/2}.$$

For any two sequences $a_n, b_n \in \mathbb{R}$, we say that $a_n \asymp b_n$ if $cb_n \leq a_n \leq Cb_n$ for some constants $c, C$.

## 2.1 Model

Let $\{\boldsymbol{X}^u\}_{u\in[0,1]}$ be a series of $d$-dimensional random vectors indexed by the label $u$, which can represent any kind of ordering in subjects (e.g., any covariate or confounder of interest transformed to the space $[0,1]$). For any $u \in [0,1]$, assume that $\boldsymbol{X}^u \sim N_d\{\boldsymbol{0}, \boldsymbol{\Sigma}(u)\}$. Here $\boldsymbol{\Sigma}(\cdot) : [0,1] \to \mathbb{S}_+^{d\times d}$ is a function from $[0,1]$ to the $d$ by $d$ positive definite matrix set, $\mathbb{S}_+^{d\times d}$. Let $\boldsymbol{\Omega}(u) := \{\boldsymbol{\Sigma}(u)\}^{-1}$ be the inverse covariance matrix of $\boldsymbol{X}^u$ and let $\mathbf{G}(u) \in \{0,1\}^{d\times d}$ represent the conditional independence graph corresponding to $\boldsymbol{X}^u$, satisfying that

$$\{\mathbf{G}(u)\}_{jk} = 1 \text{ if and only if } \{\boldsymbol{\Omega}(u)\}_{jk} \neq 0.$$

Suppose that data points in $u = u_1, \ldots, u_n$ are observed. Let $\boldsymbol{x}_{i1}, \ldots, \boldsymbol{x}_{iT} \in \mathbb{R}^d$ be $T$ observations of $\boldsymbol{X}^{u_i}$, with a temporal dependence structure among them. In particular, for simplicity, in this manuscript we assume that $\{\boldsymbol{x}_{it}\}_{t=1}^T$ follows a lag one stationary vector autoregressive (VAR) model, i.e.,

$$\boldsymbol{x}_{it} = \mathbf{A}(u_i)\boldsymbol{x}_{i(t-1)} + \boldsymbol{\epsilon}_{it}, \quad \text{for } i = 1, \ldots, n, \ t = 2, \ldots, T, \tag{2.1}$$

and $\boldsymbol{x}_{it} \sim N_d\{\boldsymbol{0}, \boldsymbol{\Sigma}(u_i)\}$ for $t = 2, \ldots, T$. Here we note that extensions to vector autoregressive models with higher orders are also analyzable using the same techniques in Han and Liu (2013). But for simplicity, in this manuscript we only consider the lag one case. $\mathbf{A}(u) \in \mathbb{R}^{d\times d}$ is referred to as the transition matrix. It is assumed that the Gaussian noise, $\boldsymbol{\epsilon}_{it} \sim N_d\{\boldsymbol{0}, \boldsymbol{\Psi}(u_i)\}$ is independent of $\{\boldsymbol{\epsilon}_{it'}\}_{t'\neq t}$ and $\{\boldsymbol{x}_{it'}\}_{t'=1}^{t-1}$. Both $\mathbf{A}(\cdot)$ and $\boldsymbol{\Psi}(\cdot)$ are considered as functions on $[0,1]$. Due to the stationary property, for any $u \in [0,1]$, taking the covariance on either side of Equation (2.1), we have

$$\boldsymbol{\Sigma}(u) = \mathbf{A}(u)\boldsymbol{\Sigma}(u)\{\mathbf{A}(u)\}^\mathsf{T} + \boldsymbol{\Psi}(u).$$

For any $i \neq i'$, it is assumed that $\{\boldsymbol{x}_{it}\}_{t=1}^T$ are independent of $\{\boldsymbol{x}_{i't}\}_{t=1}^T$. For $i = 1, \ldots, n$ and $t = 1, \ldots, T$, denote $\boldsymbol{x}_{it} = (x_{it1}, \ldots, x_{itd})^\mathsf{T}$.

Of note, the function $\mathbf{A}(\cdot)$ characterizes the temporal dependence in the time series. For each label $u$, $\mathbf{A}(u)$ represents the transition matrix of the VAR model specific to $u$. By allowing $\mathbf{A}(u)$ to depend on $u$, as $u$ varies, the temporal dependence structure of the corresponding time series is allowed to vary, too.

As is noted in Section 1, the proposed model is motivated by brain network estimation using rs-fMRI data. For instance, the ADHD data considered in Section 4.5 consist of $n$ subjects with ages ($u$) ranging from 7 to 22, while time series measurements within each subject are indexed by $t$ varying from 1 to 200, say. That is, for each subject, a list of rs-fMRI images with temporal dependence are available. We model the list of images by a VAR process, as exploited in Equation (2.1). For a fixed age $u$, $\mathbf{A}(u)$ characterizes the temporal dependence structure of the time series corresponding to the subject with age $u$. As age varies, the temporal dependence structures of the images may vary, too. Allowing $\mathbf{A}(u)$ to change with $u$ accommodates such changes. The VAR model is a common tool in modeling dependence for rs-fMRI data. Consider Harrison et al. (2003), Penny et al. (2005), Rogers et al. (2010), Chen et al. (2011), and Valdés-Sosa et al. (2005), for more details.

## 2.2 Method

We exploit the idea proposed in Zhou et al. (2010) and use a kernel based estimator for subject specific graph estimation. The proposed approach requires two main steps.. In the first step, a smoothed estimate of the covariance matrix $\mathbf{\Sigma}(u_0)$, denoted as $\mathbf{S}(u_0)$, is obtained for a target label $u_0$. In the second step, $\mathbf{\Omega}(u_0)$ is estimated by plugging the covariance matrix estimate $\mathbf{S}(u_0)$ into the CLIME algorithm (Cai et al., 2011).

More specifically, let $K(\cdot) : \mathbb{R} \to \mathbb{R}$ be a symmetric nonnegative kernel function with support set $[-1, 1]$. Moreover, for some absolute constant $C_1$, let $K(\cdot)$ satisfy that:

$$\sup_v K(v) \leq C_1, \quad \int_{-1}^{1} K(v)dv = 1, \quad \text{and} \quad \int_{0}^{1} vK(v)dv \leq C_1. \tag{2.2}$$

Equation (2.2) is satisfied by a number of commonly used kernel functions. Examples include:

Uniform kernel: $K(s) = I(|s| \leq 1)/2$;
Triangular kernel: $K(s) = (1 - |s|)I(|s| \leq 1)$;
Epanechnikov kernel: $K(s) = 3(1 - s^2)I(|s| \leq 1)/4$;
Cosine kernel: $K(s) = \pi \cos(\pi s/2)I(|s| \leq 1)/4$.

For estimating any covariance matrix $\mathbf{\Sigma}(u_0)$ with the label $u_0 \in [0, 1]$, the smoothed sample covariance matrix estimator $\mathbf{S}(u_0)$ is calculated as follows:

$$\mathbf{S}(u_0) := \sum_{i=1}^{n} \omega_i(u_0, h) \widehat{\mathbf{\Sigma}}_i, \tag{2.3}$$

where $\omega_i(u_0, h)$ is a weight function and $\widehat{\mathbf{\Sigma}}_i$ is the sample covariance matrix of $\boldsymbol{x}_{i1}, \ldots, \boldsymbol{x}_{iT}$:

$$\omega_i(u_0, h) := \frac{c(u_0)}{nh} K\left(\frac{u_i - u_0}{h}\right), \quad \widehat{\mathbf{\Sigma}}_i := \frac{1}{T} \sum_{t=1}^{T} \boldsymbol{x}_{it} \boldsymbol{x}_{it}^{\mathsf{T}} \in \mathbb{R}^{d \times d}. \tag{2.4}$$

Here $c(u_0) = 2I(u_0 \in \{0,1\}) + I\{u_0 \in (0,1)\}$ is a constant depending on whether $u_0$ is on the boundary or not, and $h$ is the bandwidth parameter. We will discuss how to select $h$ in the next section.

After obtaining the covariance matrix estimate, $\mathbf{S}(u_0)$, we proceed to estimate $\mathbf{\Omega}(u_0) := \{\mathbf{\Sigma}(u_0)\}^{-1}$. When a suitable sparsity assumption on the inverse covariance matrix $\mathbf{\Omega}(u_0)$ is available, we propose to estimate $\mathbf{\Omega}(u_0)$ by plugging $\mathbf{S}(u_0)$ into the CLIME algorithm (Cai et al., 2011). In detail, the inverse covariance matrix estimator $\widehat{\mathbf{\Omega}}(u_0)$ of $\mathbf{\Omega}(u_0)$ is calculated via solving the following optimization problem:

$$\widehat{\mathbf{\Omega}}(u_0) = \operatorname*{argmin}_{\mathbf{M} \in \mathbb{R}^{d \times d}} \sum_{jk} |\mathbf{M}_{jk}|, \quad \text{subject to } \|\mathbf{S}(u_0)\mathbf{M} - \mathbf{I}_d\|_{\max} \leq \lambda, \tag{2.5}$$

where $\mathbf{I}_d \in \mathbb{R}^{d \times d}$ is the identity matrix and $\lambda$ is a tuning parameter. Equation (2.5) can be further decomposed into $d$ optimization subproblems (Cai et al., 2011). For $j = 1, \ldots, d$, the $j$-th column of $\widehat{\mathbf{\Omega}}(u_0)$ can be solved as:

$$\{\widehat{\mathbf{\Omega}}(u_0)\}_{*j} = \operatorname*{argmin}_{\boldsymbol{v} \in \mathbb{R}^d} \|\boldsymbol{v}\|_1, \quad \text{subject to } \|\mathbf{S}(u_0)\boldsymbol{v} - \boldsymbol{e}_j\|_\infty \leq \lambda, \tag{2.6}$$

where $\boldsymbol{e}_j$ is the $j$-th canonical vector. Equation (2.6) can be solved efficiently using a parametric simplex algorithm (Pang et al., 2013). Hence, the solution to Equation (2.5) can be computed in parallel.

Once $\widehat{\mathbf{\Omega}}(u_0)$ is obtained, we can apply an additional threshold step to estimate the Graph $\mathbf{G}(u_0)$. We define a graph estimator $\widehat{\mathbf{G}} \in \{0,1\}^{d \times d}$ to be:

$$\{\widehat{\mathbf{G}}(u_0)\}_{jk} = \begin{cases} 1 & \text{if } \left|\{\widehat{\mathbf{\Omega}}(u_0)\}_{jk}\right| > \gamma, \\ 0 & \text{otherwise.} \end{cases} \tag{2.7}$$

Here $\gamma$ is another tuning parameter.

Of note, although two tuning parameters, $\lambda$ and $\gamma$, are introduced, $\gamma$ is introduced merely for theoretical soundness. Empirically, we found that setting $\gamma$ to be 0 or a very small value (e.g., $10^{-5}$) has proven to work well. This is consistent with existing literature on graphical model estimation. We refer the readers to Cai et al. (2011), Liu et al. (2012a), Liu et al. (2012b), Xue and Zou (2012), and Han et al. (2013) for more discussion on this issue.

Procedures for choosing $\lambda$ have also been well studied in the graphical model literature. On one hand, popular selection criteria, such as the stability approach based on subsampling (Meinshausen and Bühlmann, 2010; Liu et al., 2010), exist and have been well studied. On the other hand, when prior knowledge about the sparsity of the precision matrix is available, a common approach is trying a sequence of $\lambda$, and choosing one according to a desired sparsity level.

## 3 Theoretical Properties

In this section the theoretical properties of the proposed estimators in Equations (2.5) and (2.7) are provided. Under a double asymptotic framework, the rates of convergence in parameter estimation under the matrix $\ell_1$ and $\ell_{\max}$ norms are given.

Before establishing the theoretical result, we first pose an additional assumption on the function $\mathbf{\Sigma}(\cdot)$. In detail, let $\Sigma_{jk}(\cdot) : u \to \{\mathbf{\Sigma}(u)\}_{jk}$ be a real function. In the following, we assume that $\Sigma_{jk}(\cdot)$ is a smooth function with regard to any $j, k \in \{1, \ldots, d\}$. Here and in the sequel, the derivatives at support boundaries are defined as one-sided derivatives.

**(A1)** There exists one absolute constant, $C_2$, such that for all $u \in [0,1]$,

$$\left| \frac{d}{du} \Sigma_{jk}(u) \right| \leq C_2, \quad \text{for } j, k \in \{1, \ldots, d\}.$$

Under Assumption **(A1)**, we propose the following lemma, which shows that when the subjects are sampled in $u = u_1, \ldots, u_n$ with $u_i = i/n$ for $i = 1, \ldots, n$, the estimator $\mathbf{S}(u_0)$ approximates $\mathbf{\Sigma}(u_0)$ at a fast rate for any $u_0 \in [0,1]$. The convergence rate delivered here characterizes both the strength one can borrow across different subjects and the impact of temporal dependence structure on estimation accuracy.

**Lemma 3.1.** *Suppose that the data points are generated from the model discussed in Section 2.1 and Assumption **(A1)** holds. Moreover, suppose that the observed subjects are in $u_i = i/n$ for $i = 1, \ldots, n$. Then, for any $u_0 \in [0,1]$, if for some $\eta > 0$ we have*

$$(\mathbf{A2}) \quad \sup_{u \in [0,1]} \frac{d^2}{du^2} \left\{ K\left(\frac{u - u_0}{h}\right) \Sigma_{jk}(u) \right\} = O(h^{-\eta}), \quad \text{for } j, k \in \{1, \ldots, d\},$$

*and the bandwidth $h$ is set as*

$$h \asymp \max\left\{ \left\{ \frac{\xi \cdot \sup_{u \in [0,1]} \|\mathbf{\Sigma}(u)\|_2}{1 - \sup_{u \in [0,1]} \|\mathbf{A}(u)\|_2} \sqrt{\frac{\log d}{Tn}} \right\}^{1/2}, n^{-\frac{2}{2+\eta}} \right\}, \quad (3.1)$$

*where*

$$\xi := \sup_{u \in [0,1]} \frac{\max_j [\mathbf{\Sigma}(u)]_{jj}}{\min_j [\mathbf{\Sigma}(u)]_{jj}},$$

then the smoothed sample covariance matrix estimator $\mathbf{S}(u_0)$ defined in Equation (2.3) satisfies:

$$\|\mathbf{S}(u_0) - \mathbf{\Sigma}(u_0)\|_{\max} = O_P\left[\left\{\frac{\xi \sup_{u\in[0,1]} \|\mathbf{\Sigma}(u)\|_2}{1 - \sup_{u\in[0,1]} \|\mathbf{A}(u)\|_2}\sqrt{\frac{\log d}{Tn}}\right\}^{1/2} + n^{-\frac{2}{2+\eta}}\right]. \quad (3.2)$$

Assumption **(A2)** is a convolution between the smoothness of $K(\cdot)$ and $\Sigma_{jk}(\cdot)$, and is a weaker requirement than imposing smoothness individually. Assumption **(A2)** is satisfied by many commonly used kernel functions, including the aforementioned examples in Section 2.2. For example, with regard to the Epanechnikov kernel $K(s) = 3(1 - s^2)I(|s| \leq 1)/4$, it's easy to check that

$$\frac{d}{du}K\left(\frac{u - u_0}{h}\right) = O\left(\frac{1}{h^2}\right) \quad \text{and} \quad \frac{d^2}{du^2}K\left(\frac{u - u_0}{h}\right) = O\left(\frac{1}{h^2}\right).$$

Therefore, as long as $\Sigma_{jk}(u)$, $\frac{d}{du}\Sigma_{jk}(u)$, and $\frac{d^2}{du^2}\Sigma_{jk}(u)$ are uniformly bounded, the Epanechnikov kernel satisfies Assumption **(A2)** with $\eta \geq 2$.

There are several observations drawn from Lemma 3.1. First, the rate of convergence in parameter estimation is upper bounded by $n^{-\frac{2}{2+\eta}}$, which is due to the bias in estimating $\mathbf{\Sigma}(u_0)$ from only $n$ labels. This term is irrelevant to the sample size $T$ in each subject and cannot be improved without adding stronger (potentially unrealistic) assumptions. For example, when none of $\xi$, $\sup_t \|\mathbf{\Sigma}(u)\|_2$, and $\sup_t \|\mathbf{A}(u)\|_2$ scales with $(n, T, d)$ and $T > Cn^{\frac{6-\eta}{2+\eta}} \log d$ for some generic constant $C$, the estimator achieves a $n^{-\frac{2}{2+\eta}}$ rate of convergence. Secondly, in the term $\{\log d/(Tn)\}^{1/4}$, $n$ characterizes the strength one can borrow across different subjects, while $T$ demonstrates the contribution from within a subject. When $n > CT^{\frac{2+\eta}{6-\eta}}$, the estimator achieves a $\{\log d/(Tn)\}^{1/4}$ rate of convergence. The first two points discussed above, together, quantify the settings where the proposed methods can beat the naive method which only exploits the data points in each subject itself for parameter estimation.

Finally, Lemma 3.1 also demonstrates how temporal dependence may affect the rate of convergence. Specifically, the spectral norm of the transition matrix, $\|\mathbf{A}(u)\|_2$, characterizes the strength of temporal dependence. The term $1/\{1-\sup_{u\in[0,1]}\|\mathbf{A}(u)\|_2\}$ in Equation (3.2) demonstrates the impact of the dependence strength on the rate of convergence.

Next we investigate the effect of the sign and strength of auto-correlation and cross-correlation on the rate of convergence. In detail, we define the diagonal entries of $\mathbf{A}(u)$ to be the auto-correlation coefficients, since they capture how $(\boldsymbol{x}_{it})_j$ depends on $\{\boldsymbol{x}_{i(t-1)}\}_j$, for $i = 1, \ldots, n$, $t = 2, \ldots, T$, and $j = 1, \ldots, d$. We define the off-diagonal entries of $\mathbf{A}(u)$ to be the cross-correlation coefficients, since they capture how $(\boldsymbol{x}_{it})_j$ depends on $\{\boldsymbol{x}_{i(t-1)}\}_{\setminus j}$.

Since a general analysis is intractable, we focus on several special structures on $\mathbf{A}(u)$. We suppress the label $u$ in $\mathbf{A}(u)$, and subject index $i$ in $\boldsymbol{x}_{it}$ for notational brevity.

1. We first study the effect of auto-correlation. For highlighting autocorrelation alone, we set the cross-correlation coefficients to be 0 and consider the case where $\mathbf{A}$ is diagonal: $\mathbf{A} = \mathrm{diag}(\rho_1, \ldots, \rho_d)$. This scenario is equivalent to $d$ independent time series.

2. Secondly, we study the effect of the cross-correlation. To this end, we set the diagonal entries of $\mathbf{A}$ to be 0. In this scenario, at any time point, a variable does not depend on its value at the previous time point in the autoregression. Below we focus on two special structures on the off-diagonal entries, as exploited in Han and Liu (2013).

    (a) $\mathbf{A}$ has a "band" structure, i.e., $\mathbf{A}_{ij} = \rho I(|i-j|=1)$. In this case, the $j$-th entry of $\boldsymbol{x}_t$ only depends on adjacent entries at time $t-1$, i.e., entries in $\boldsymbol{x}_{t-1}$ with index differing from $j$ by 1.

    (b) $\mathbf{A}$ is block diagonal. Each block has an "AR" structure. Specifically, let $\mathbf{A} = \mathrm{diag}(\mathbf{A}_1, \ldots, \mathbf{A}_k)$, where $\mathbf{A}_l \in \mathbb{R}^{d_l \times d_l}$ for $l = 1, \ldots, k$. We have $(\mathbf{A}_l)_{ij} = \rho^{|i-j|} I(i \neq j)$, for $i, j = 1, \ldots, d_l$. In this case, the entries of $\boldsymbol{x}_t$ form $k$ clusters. Temporal dependence occurs only within clusters. In each cluster, the cross-correlation coefficients decrease exponentially with the gap in index.

The next theorem summarizes the impact of the correlation coefficients on the rate of convergence.

**Theorem 3.2.** *Let $\mathbf{A}$ be one of the transition matrices defined in (1), (2).i and (2).ii. Inheriting the assumptions and notations in Lemma 3.1, we have:*

*(1). Under Scenario (1), we have*

$$\|\mathbf{S}(u_0) - \boldsymbol{\Sigma}(u_0)\|_{\max} = O_P \left[ \left\{ \frac{\xi \sup_{u \in [0,1]} \|\boldsymbol{\Sigma}(u)\|_2}{1 - \max_{j=1,\ldots,d}(|\rho_j|)} \sqrt{\frac{\log d}{Tn}} \right\}^{1/2} + n^{-\frac{2}{2+\eta}} \right].$$

*Thus, the magnitude of the maximum auto-correlation coefficient has a negative effect on the convergence rate. In comparison, the signs of the auto-correlation coefficients has no effect.*

*(2). Under Scenario (2). i, we have*

$$\|\mathbf{S}(u_0) - \boldsymbol{\Sigma}(u_0)\|_{\max} = O_P \left[ \left\{ \frac{\xi \sup_{u \in [0,1]} \|\boldsymbol{\Sigma}(u)\|_2}{1 - 2|\rho| \cos\{\pi/(d+1)\})} \sqrt{\frac{\log d}{Tn}} \right\}^{1/2} + n^{-\frac{2}{2+\eta}} \right].$$

*Under Scenario (2).ii, we have $\|\mathbf{S}(u_0) - \mathbf{\Sigma}(u_0)\|_{\max} = O_P[\alpha(\rho, \xi, \mathbf{\Sigma}, T, n, d)]$, where $\alpha$, as a function of $\rho$, is symmetric around $0$ and monotonically increasing in for $\rho > 0$. Thus, the magnitude of the cross-correlation coefficients has a negative effect on the convergence rate. Again, the signs of the cross-correlation coefficients has no effect.*

Although Theorem 3.2 only presents the effect of the correlation coefficients on the upper bound of estimation error, the simulation study in Section 4.2 provides consistent results in estimation accuracy.

Next, we consider the case where $\mathbf{A}(u) = 0$ and hence $\{\boldsymbol{x}_{it}\}_{t=1}^T$ are independent observations with no temporal dependence. In this case, following Zhou et al. (2010), the rate of convergence in parameter estimation for the proposed approach can be improved.

**Lemma 3.3.** *Under the assumptions in Lemma 3.1, if it is further assumed that*

**(B1)** $\{\boldsymbol{x}_{it}\}_{t=1}^T$ *are i.i.d. observations from $N_d\{\mathbf{0}, \mathbf{\Sigma}(u)\}$;*

**(B2)** $\sup_{u \in [0,1]} \frac{d^2}{du^2} \left[ K^2\left(\frac{u-u_0}{h}\right) \left\{ \Sigma_{jj}^2(u)\Sigma_{kk}^2(u) + \Sigma_{jk}^2(u) \right\} \right] = O(h^{-4})$ *for all $j, k \in \{1, \ldots, d\}$;*

**(B3)** *There exists an absolute constant $C_3$ such that*

$$\max_{jk} \sup_{u \in [0,1]} |\Sigma_{jk}(u)| \leq C_3, \quad \max_{jk} \sup_{u \in [0,1]} \left| \frac{d}{du} \Sigma_{jk}(u) \right| \leq C_3;$$

*then, setting the bandwidth*

$$h \asymp \max\left\{ \left(\frac{\log d}{Tn}\right)^{1/3}, \frac{1}{n^{2/(2+\eta)}} \right\}, \tag{3.3}$$

*we have*

$$\|\mathbf{S}(u_0) - \mathbf{\Sigma}(u_0)\|_{\max} = O_P\left\{ \left(\frac{\log d}{Tn}\right)^{1/3} + n^{-\frac{2}{2+\eta}} \right\}.$$

We note again that the aforementioned kernel functions satisfy Assumptions **(B2)** for similar reasons. In detail, taking Epanechnikov kernel as an example, we have

$$\frac{d}{du} K^2\left(\frac{u-u_0}{h}\right) = O\left(\frac{1}{h^4}\right), \quad \frac{d^2}{du^2} K^2\left(\frac{u-u_0}{h}\right) = O\left(\frac{1}{h^4}\right).$$

So Assumption **(B2)** is satisfied as long as $\Sigma_{jk}(u)$, $\frac{d}{du}\Sigma_{jk}(u)$, and $\frac{d^2}{du^2}\Sigma_{jk}(u)$ are uniformly bounded.

Lemma 3.3 shows that the rate of convergence can be improved to $\{\log d/(Tn)\}^{1/3}$ when the data are independent. Of note, this rate matches the results in Zhou et al. (2010). However, the improved rate is valid only when a strong independence assumption holds, which is unrealistic in many applications, rs-fMRI data analysis for example.

After obtaining Lemmas 3.1 and 3.3, we proceed to the final result, which shows the theoretical performance of the estimators $\widehat{\boldsymbol{\Omega}}(u_0)$ and $\widehat{\mathbf{G}}(u_0)$ proposed in Equations (2.5) and (2.7). We show that under certain sparsity constraints, the proposed estimators are consistent, even when $d$ is nearly exponentially larger than $n$ and $T$.

We first introduce some additional notation. Let $M_d \in \mathbb{R}$ be a quantity which may scale with $(n, T, d)$. We define the set of positive definite matrices in $\mathbb{R}^{d \times d}$, denoted by $\mathcal{M}(q, s, M_d)$, as

$$\mathcal{M}(q, s, M_d) := \left\{ \mathbf{M} \in \mathbb{R}^{d \times d} : \max_{1 \leq k \leq d} \sum_{j=1}^{d} |M_{jk}|^q \leq s, \|\mathbf{M}\|_1 \leq M_d \right\}.$$

For $q = 0$, the class $\mathcal{M}(0, s, M_d)$ contains all the matrices with the number of nonzero entries in each column less than $s$ and bounded $\ell_1$ norm. We then let

$$\kappa(n, T, d) := \left\{ \frac{\xi \sup_{u \in [0,1]} \|\boldsymbol{\Sigma}(u)\|_2}{1 - \sup_{u \in [0,1]} \|\mathbf{A}(u)\|_2} \sqrt{\frac{\log d}{Tn}} \right\}^{1/2} + n^{-\frac{2}{2+\eta}}, \tag{3.4}$$

$$\kappa^*(n, T, d) := \left( \frac{\log d}{Tn} \right)^{1/3} + n^{-\frac{2}{2+\eta}}. \tag{3.5}$$

Theorem 3.4 presents the parameter estimation and graph estimation consistency results for the estimators defined in Equations (2.5) and (2.7).

**Theorem 3.4.** *Suppose that the conditions in Lemma 3.1 hold. Assume that $\boldsymbol{\Theta}(u_0) := \{\boldsymbol{\Sigma}(u_0)\}^{-1} \in \mathcal{M}(q, s, M_d)$ with $0 \leq q < 1$. Let $\widehat{\boldsymbol{\Theta}}(u_0)$ be defined in Equation (2.5). Then there exists a constant $C_3$ only depending on $q$, such that, whenever the tuning parameter*

$$\lambda = C_3 M_d \kappa(n, T, d)$$

*is chosen, one has that*

$$\|\widehat{\boldsymbol{\Theta}}(u_0) - \boldsymbol{\Theta}(u_0)\|_2 = O_P \left\{ M_d^{2-2q} s \kappa(n, T, d)^{1-q} \right\}.$$

*Moreover, let $\widehat{\mathbf{G}}(u_0)$ be the graph estimator defined in Equation (2.7) with the second step tuning parameter $\gamma = 4 M_d \lambda$. If it is further assumed that $\boldsymbol{\Theta}(u_0) \in \mathcal{M}(0, s, M_d)$ and*

$$\min_{\{j,k : |\{\boldsymbol{\Theta}(u_0)\}_{jk}| \neq 0\}} |\{\boldsymbol{\Theta}(u_0)\}_{jk}| \geq 2\gamma,$$

*then*

$$\mathbb{P}\left\{\widehat{\mathbf{G}}(u_0) = \mathbf{G}(u_0)\right\} = 1 - o(1).$$

*If the conditions in Lemma 3.3 hold, the above results are true with $\kappa$ replaced by $\kappa^*$.*

Theorem 3.4 shows that the proposed method is theoretically guaranteed to be consistent in both parameter estimation and model selection, even when the dimension $d$ is nearly exponentially larger than $nT$. Theorem 3.4 can be proved by following the proofs of Theorem 1 and Theorem 7 in Cai et al. (2011) and the proof is accordingly omitted.

# 4 Experiments

In this section, the empirical performance of the proposed method is investigated. This section consists of two parts. In the first, we demonstrate the performance using synthetic data, where the true generating models are known. On one hand, the proposed kernel based method is compared to several existing methods. The advantage of this new method is shown in both parameter estimation and model selection. On the other hand, implications of the theoretical results in Section 3 are also empirically verified. In the second part, the proposed method is applied to a large scale rs-fMRI data (the ADHD-200 data) and some potentially scientifically interesting results are explored.

## 4.1 Synthetic Data

The performance of the proposed kernel-smoothing estimator (denoted as KSE) is compared to three existing methods: a naive estimator (donated as naive; details follow below), Danaher et al. (2014)'s group graphical lasso (denoted as GGL), and Guo et al. (2011)'s estimator (denoted as Guo). Throughout the simulation studies, it is assumed that the graphs are evolving from $u = 0$ to $u = 1$ continuously. Although there is one graphical model corresponding to each $u \in [0, 1]$, it is assumed that data are observed at $n$ equally spaced points $u = 0, 1/(n-1), 2/(n-1), \ldots, 1$. For each $u = 0, 1/(n-1), 2/(n-1), \ldots, 1$, $T$ observations were generated from the corresponding graph under a stationary VAR(1) model discussed in Equation (2.1). To generate the transition matrix, $\mathbf{A}$, the precision matrix was obtained using the R package *Huge* (Zhao et al., 2012) with graph structure "random". Then it is divided by twice its largest eigenvalue to obtain $\mathbf{A}$, so that $\|\mathbf{A}\|_2 = 0.5$. The same transition matrix is used under every label $u$. Our main target is to estimate the graph at $u_0 = 0$, as the endpoints represent the most difficult point for estimation.

We also investigate one setting where the target label is $u_0 = 1/2$, to demonstrate the performance at a non-extreme target label.

In the following, three existing methods for comparison are reviewed. naive is obtained by first calculating the sample covariance matrix at target label $u_0$ using only the $T$ observations under this label, and then plugged into the CLIME algorithm. Compared to KSE, GGL and Guo do not assume that there exists a smooth change among the graphs. Instead, they assume that the data come from $n$ categories. That is, there are $n$ corresponding underlying graphs that potentially share common edges, and observations are available within each category. Moreover, they assume that the observations are independent both between and within different categories. With regard to implementation, they solve the following optimization problem:

$$\max_{\boldsymbol{\Omega}^{(0)},\ldots,\boldsymbol{\Omega}^{(n)} \succ 0} \sum_{i=0}^{n} T \left\{ \log \det \boldsymbol{\Omega}^{(i)} - \operatorname{trace}\left(\widehat{\boldsymbol{\Sigma}}_i \boldsymbol{\Omega}^{(i)}\right) \right\} - P\left(\boldsymbol{\Omega}^{(0)}, \ldots, \boldsymbol{\Omega}^{(n)}\right),$$

where $\widehat{\boldsymbol{\Sigma}}_i$ is the sample covariance matrix calculated based on the data under label $u_i$. GGL uses penalty

$$P\left(\boldsymbol{\Omega}^{(0)}, \ldots, \boldsymbol{\Omega}^{(n)}\right) = \lambda_1 \sum_{i=0}^{n} \sum_{j \neq k} |\{\boldsymbol{\Omega}^{(i)}\}_{jk}| + \lambda_2 \sum_{j \neq k} \sqrt{\sum_{i=0}^{n} \{\boldsymbol{\Omega}^{(i)}\}_{jk}^2},$$

and Guo uses penalty

$$P\left(\boldsymbol{\Omega}^{(0)}, \ldots, \boldsymbol{\Omega}^{(n)}\right) = \lambda \sum_{j \neq k} \sqrt{\sum_{i=0}^{n} |\{\boldsymbol{\Omega}^{(i)}\}_{jk}|}.$$

Here the regularity coefficients $\lambda_1$, $\lambda_2$, and $\lambda$ control the sparsity level. Danaher et al. (2014) also proposed the fused graphical lasso that separately controls sparsity of and similarity between the graphs. However, this method is not scalable when the number of categories is large and therefore not included in our comparison.

After obtaining the estimated graph, $\widehat{\mathbf{G}}(u_0)$, of the true traph $\mathbf{G}(u_0)$, the model selection performance is further investigated by comparing the ROC curves of the four competing methods. Let $\widehat{\mathrm{E}}(u_0)$ be the set of estimated edges corresponding to $\widehat{\mathbf{G}}(u_0)$, and $\mathrm{E}(u_0)$ the set of true edges corresponding to $\mathbf{G}(u_0)$. The true positive rate (TPR) and false positive rate (FPR) are defined as

$$\mathrm{TPR} = \frac{|\widehat{\mathrm{E}}(u_0) \cap \mathrm{E}(u_0)|}{|\mathrm{E}(u_0)|}, \quad \mathrm{FPR} = \frac{|\widehat{\mathrm{E}}(u_0) \setminus \mathrm{E}(u_0)|}{d(d-1)/2 - |\mathrm{E}(u_0)|},$$

where for any set S, |S| denotes the cardinality of S. To obtain a series of TPRs and FPRs, for KSE, naive, and Guo, the values of $\lambda$ are varied. For GGL, first $\lambda_2$ is fixed and subsequently $\lambda_1$ is tuned, and then the $\lambda_2$ with the best overall performance is selected. More specifically, a series of $\lambda_2$ are picked, and for each fixed $\lambda_2$, $\lambda_1$ is accordingly varied to produce an ROC curve. Of note, in the investigation, the ROC curves indexed by $\lambda_2$ are generally parallel, thus motivating this strategy. Finally, the $\lambda_2$ corresponding to the topleft most curve is selected.

### 4.1.1 Setting 1: Simultaneously Evolving Edges

In this section we investigate the performance of the four competing methods under one particular graphical model. In each simulation, $n_{\text{fix}} = 200$ edges are randomly selected from $d(d-1)/2$ potential edges and they do not change with regard to the label $u$. The strengths of these edges, i.e. the corresponding entries in the inverse covariance matrix, are generated from a uniform distribution taking values in $[-0.3, -0.1]$ (denoted by Unif$[-0.3, -0.1]$) and do not change with $u$. We then randomly select $n_{\text{decay}}$ and $n_{\text{grow}}$ edges that will disappear and emerge over the evolution simultaneously. For each of the $n_{\text{decay}}$ edges, the strength is generated from Unif[-0.3,-0.1] at $u = 0$ and will diminish to 0 linearly with regard to $u$. For each of the $n_{\text{grow}}$ edges, the strength is set to be 0 at $u = 0$, and will linearly grow to a value generated from Unif[-0.3,-0.1]. The edges evolve simultaneously. For $j \neq k$, when we subtract a value $a$ from $\mathbf{\Omega}_{jk}$ and $\mathbf{\Omega}_{kj}$, we increase $\mathbf{\Omega}_{jj}$ and $\mathbf{\Omega}_{kk}$ by $a$, and then further add 0.25 to the diagonal of the matrix to keep it positive definite.

The ROC curves under this setting with different values of $n_{\text{grow}}$ and $n_{\text{decay}}$ are shown in Figures 1(a) and 1(b). We fix the number of labels $n = 51$, number of observations under each label $T = 100$, and dimension $d = 50$. The target label is $u_0 = 0$. It can be observed that, under both cases, KSE outperforms the other three competing methods. Moreover, when we increase the values of $n_{\text{grow}}$ and $n_{\text{decay}}$ from 20 to 100, the ROC curve of KSE hardly changes, since the degree of smoothness in graphical model evolving hardly change. In contrast, the ROC curves of GGL and Guo drop, since the degree of similarity among the graphs is reduced. Finally, naive performances worst, which is expected because it does not borrow strength across labels in estimation. Figure 1(c) illustrates the performance under the same setting as in Figure 1(a) except $u_0 = 1/2$. KSE still outperforms the other estimators.

Next, we exploit the same data, but permute the labels $u = 1/50, 2/50, \ldots, 1$ so that the evolving pattern is much more opaque. Figures 1(d) and 1(e) illustrate the model selection result. We observe that under this setting, the ROC curves of the proposed method drop a little bit, but is still higher than the competing approaches. This is because the proposed method still benefits from the evolving graph structure (although more turbulent this time). The improvement over the naive method demonstrates exactly the strength borrowed across different labels. Note that the ROC curves of GGL, naive, and Guo shown in Figures 1(d) and 1(e) do not change compared to those in Figures 1(a) and 1(b), respectively, because they do not assume any ordering between the graphs.

### 4.1.2 Setting 2: Sequentially Growing Edges

Setting 2 is similar to Setting 1. The two differences are: (i) Here $n_{\text{decay}}$ is set to be zero; (ii) The $n_{\text{grow}}$ edges emerges sequentially instead of simultaneously. These $n_{\text{grow}}$ edges are randomly selected, but there is no overlap with the existing 200 pre-fixed edges. The entries of the inverse covariance matrix for the $n_{\text{grow}}$ edges each grow to a value generated from Unif$[-0.3, -0.1]$, linearly in a length $1/n_{\text{grow}}$ interval in $[0, 1]$, one after another. We note that there is possibility that $n < n_{\text{grow}}$, because $n$ represents only the number of labels we observe. Under this setting, Figures 1(f) and 1(g) plot the ROC curves of the four competing methods. We also apply the four methods to the setting where the same permutation as in Setting 1 is exploited. We show the results in Figures 1(h) and 1(i). Here the same observations persist as in Setting 1.

### 4.1.3 Setting 3: Random Edges

In this setting, in contrast to the above two settings, we violate the smoothness assumption of KSE to the extreme. We demonstrate the limitedness of the proposed method in this setting. More specifically, in this setting, under every label $u$, $n_{\text{ed}}$ edges are random selected with strengths from Unif$[-0.3, -0.1]$. In this case, the graphs do not evolve smoothly over the label $u$, and the data under the labels $u \neq 0$ only contribute noises. We then apply the four competing methods to this setting and Figure 1(j) illustrates the result. Under this setting, we observe that naive beats all the other three methods. It is expected because naive is the only method that do not suffer from the noises. Here KSE performs worse than GGL and Guo, because there does not exist a natural ordering among the graphs.

Under the above three data generating settings, we further quantitatively compare the performance in parameter estimation of the inverse covariance matrix $\mathbf{\Omega}(u_0)$ for the four competing methods. Here the distances between the estimated and the true concentration matrices with regard to the matrix $\ell_1, \ell_2$, and Frobenius norms are shown in Table 1. It can be observed that KSE achieves the lowest estimation error in all settings except for the

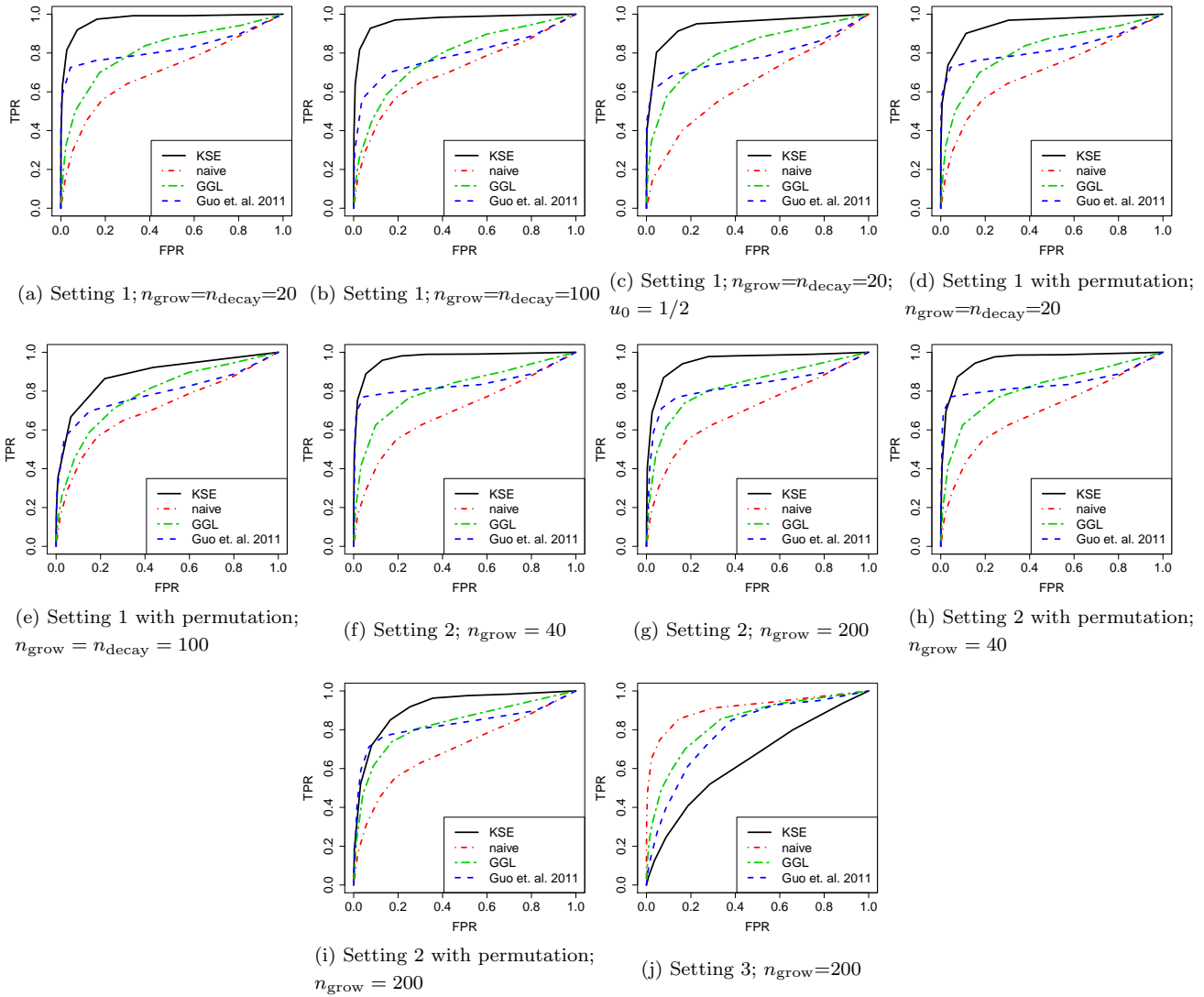

Figure 1: ROC curves of four competing methods under three settings: simultaneous (a-e), sequential (f-i), and random (j). The target labels are $u_0 = 0$ except for in (c), where $u_0 = 1/2$. In each setting we set the dimension $d = 50$, the number of labels $n = 51$, the number of observations $T = 100$, and the result is obtained by 1,000 simulations.

Setting 3. This coincides with the above model selection results. We omit the results for the label permutation cases and the case with $u_0 = 1/2$, since they are again as expected from the model selection results above.

Table 1: Comparison of inverse covariance matrix estimation errors in there data generating models. The parameter estimation error with regard to the matrix $\ell_1, \ell_2$, and Frobenius norms (denoted as $\ell_F$ here) is provided with standard deviations provided in the brackets. The results are obtained by 1,000 simulations.

|  |  | KSE | | | naive | | |
|---|---|---|---|---|---|---|---|
|  | $n_{\text{grow}} = n_{\text{decay}}$ | $\ell_1$ | $\ell_2$ | $\ell_F$ | $\ell_1$ | $\ell_2$ | $\ell_F$ |
| Setting 1 | 20 | 3.25(0.232) | 1.53(0.104) | 4.42(0.220) | 5.02(0.287) | 2.68(0.132) | 8.30(0.412) |
|  | 100 | 2.72(0.165) | 1.30(0.088) | 3.78(0.204) | 4.85(0.467) | 2.55(0.117) | 8.13(0.453) |
|  | $n_{\text{grow}}$ | | | | | | |
| Setting 2 | 40 | 3.39(0.553) | 1.56(0.213) | 4.47(0.302) | 5.26(0.740) | 2.73(0.313) | 8.24(0.386) |
|  | 200 | 3.40(0.507) | 1.57(0.147) | 4.33(0.284) | 5.19(0.740) | 2.71(0.280) | 8.34(0.352) |
| Setting 3 | $n_{\text{ed}}$ | | | | | | |
|  | 50 | 2.21(0.194) | 1.37(0.120) | 3.20(0.104) | 1.60(0.249) | 0.84(0.113) | 3.09(0.185) |
|  |  | GGL | | | Guo et. al. 2011 | | |
|  | $n_{\text{grow}} = n_{\text{decay}}$ | $\ell_1$ | $\ell_2$ | $\ell_F$ | $\ell_1$ | $\ell_2$ | $\ell_F$ |
| Setting 1 | 20 | 3.28(0.298) | 1.45(0.112) | 4.13(0.190) | 3.22(0.418) | 1.42(0.259) | 4.04(0.280) |
|  | 100 | 3.27(0.324) | 1.42(0.100) | 4.18(0.222) | 3.38(0.474) | 1.41(0.169) | 4.31(0.335) |
|  | $n_{\text{grow}}$ | | | | | | |
| Setting 2 | 40 | 3.47(0.580) | 1.47(0.163) | 4.22(0.153) | 3.06(0.417) | 1.40(0.274) | 4.00(0.205) |
|  | 200 | 3.22(0.618) | 1.44(0.198) | 4.08(0.199) | 3.71(0.493) | 1.73(0.264) | 4.46(0.361) |
| Setting 3 | $n_{\text{ed}}$ | | | | | | |
|  | 50 | 1.52(0.224) | 0.85(0.105) | 2.04(0.104) | 1.48(0.263) | 0.67(0.116) | 1.81(0.150) |

## 4.2 Impact of Temporal Dependence

In this section, we investigate the impact of temporal dependence on graph estimation accuracy. Corresponding to the discussions in Section 3, we consider three special structures of the transition matrix $\mathbf{A}(u) \in \mathbb{R}^{d \times d}$ to demonstrate the impact of auto-correlation and cross-correlation. To be illustrative, we fix the dimension $d = 10$. For simplicity, we let $\mathbf{A}(u)$ be constant over $u \in [0, 1]$, and suppress the label $u$ in $\mathbf{A}(u)$.

1. diagonal: $\mathbf{A} = \text{diag}(\rho, \ldots, \rho)$;

2. band: $\mathbf{A}_{ij} = \rho I(|i - j| = 1)$;

3. block diagonal: $\mathbf{A} = \text{diag}(\mathbf{A}_1, \mathbf{A}_2, \mathbf{A}_3)$, where $\mathbf{A}_1, \mathbf{A}_2 \in \mathbb{R}^{3 \times 3}$, and $\mathbf{A}_3 \in \mathbb{R}^{4 \times 4}$, and $(\mathbf{A}_l)_{ij} = \rho^{|i-j|} I(i \neq j)$, for $l = 1, 2, 3$.

Using these transition matrices, we generated data according to Setting 1 described in Section 4.1.1. We fixed $n = 51$, $T = 50$, and $d = 10$, and target at label $u_0 = 0$. To investigate the impact of strong versus weak auto-correlation, we range $\rho$ in $\{0.2, 0.4, 0.6\}$ and $\{-0.2, -0.4, -0.6\}$ under Scenario (1). Figures 2(a) and 2(b) display the results. One can see that large values of $|\rho|$ correspond to low estimation accuracy. Comparing Figures 2(a) and 2(b), it can be seen that the sign of the auto-correlation coefficients does not noticeably affect the ROC curves.

To investigate the impact of strong versus weak positive cross-correlation, we vary $\rho$ in $\{0.1, 0.5, 0.6\}$ under Scenarios (2) and (3). To keep $\|\mathbf{A}\|_2 < 1$, we scale $\mathbf{A}$ by $0.95/\|\mathbf{A}_{\max}\|_2$, where $\mathbf{A}_{\max}$ is the transition matrix when $\rho = 0.6$. Figures 2(c) and 2(d) show the results. Again, larger correlation results in decreased estimation accuracy.

Finally, to investigate the impact of strong positive versus strong negative cross-correlation, we compare $\rho = 0.6$ with $\rho = -0.6$ under Scenarios (2) and (3). Figures 2(e) and 2(f) deliver the results. Still the sign of cross-correlation does not dramatically affect the performance.

## 4.3 Impact of Label Size $n$, Sample Size $T$, and Dimension $d$.

In this section, we empirically demonstrate how the label size $n$, sample size $T$, and dimension $d$ may affect estimation accuracy. We inherit Setting 1 described in Section 4.1.1. We range $n$ in $\{10, 20, 40, 80\}$, $T$ in $\{25, 50, 100, 200\}$, and $d$ in $\{25, 50, 75, 100\}$. Note that when $d$ varies, $n_{\text{fix}}$, $n_{\text{grow}}$, and $n_{\text{decay}}$ are scaled to maintain the same sparsity. Figure 3 shows the results. As indicated by the rate of convergence in Section 3, estimation accuracy drops as we decrease $n$ or $T$, or increase $d$.

The simulation results in Sections 4.2 and 4.3 provide empirical support for Theorem 3.4. Although only an upper bound on the estimation error is presented in Theorem 3.4, the rate of convergence does provide informative guidance on how the parameters may affect estimation accuracy.

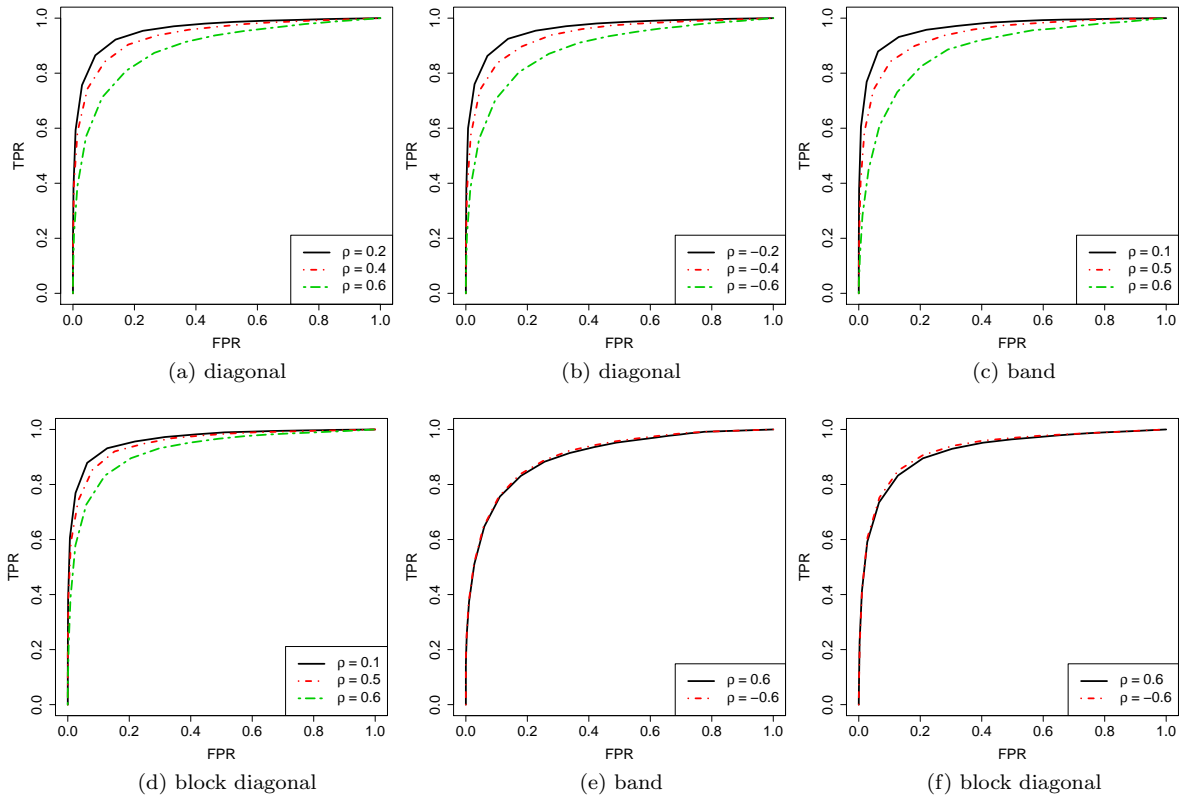

Figure 2: ROC curves of KSE under three structured transition matrices: diagonal, band, and block diagonal. Data are synthesized under Setting 1. We set dimension $d = 10$; number of labels $n = 51$; number of observations $T = 50$.

## 4.4 Impact of a Small Label Size $n$

As is shown in Lemma 3.1 and Theorem 3.4, the rates of convergence in parameter estimation and model selection crucially depend on the term $n^{-\frac{2}{2+\eta}}$. This is due to the bias in estimating $\mathbf{\Sigma}(u_0)$ from $n$ labels. This bias takes place as long as we include data under other labels into estimation, and cannot be removed by simply increasing the number of observations $T$ under each label $u$. More specifically, in the appendix, Lemma A.1 shows quantitatively that the rate of convergence for bias between the estimated and the true covariance matrix depends on $n$ but not $T$.

This section is devoted to illustrate this phenomenon empirically. We exploit Setting

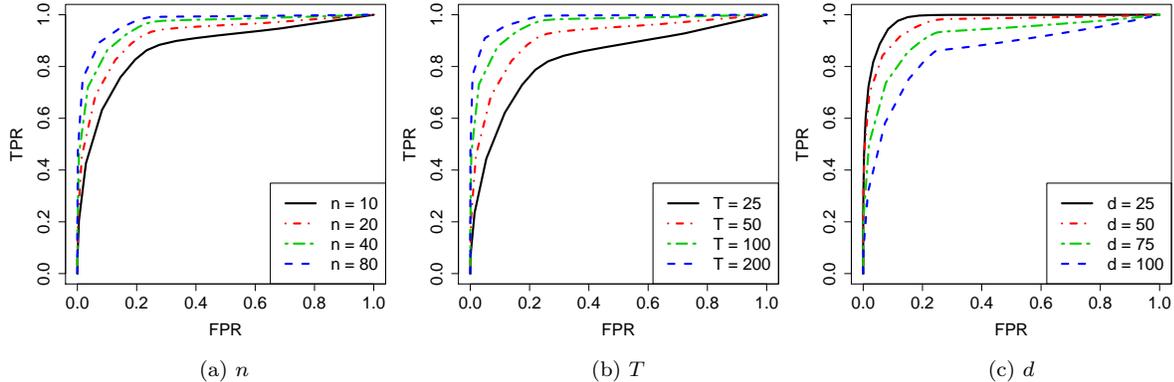

(a) $n$  (b) $T$  (c) $d$

Figure 3: ROC curves of KSE under Setting 1 with varying label size $n$, sample size $T$, or dimension $d$.

2 in the last section with the number of labels $n$ to be very small. Here we set $n = 3$. Moreover, we choose $n_{\text{fix}} = 100$, $n_{\text{grow}} = 500$, and vary the number of observations $T$ under each label. Figure 4 compares the ROC curves of KSE and naive corresponding to the settings when $T = 100$ or $500$. There are two important observations we would like to emphasize: (i) When $T = 100$, KSE and naive have comparable performance. However, when $T = 500$, naive performs much better than KSE. (ii) The change of the ROC curves for KSE between $T = 100$ and $T = 500$ is not that dramatic compared to the ROC curves for naive. These observations indicate the existence of bias in KSE that cannot be eliminated by only increasing $T$.

## 4.5 ADHD-200 Data

As an example of real data application, we apply the proposed method to the ADHD-200 data (Biswal et al., 2010). The ADHD-200 data consist of rs-fMRI images of 973 subjects. Of them, 491 are healthy and 197 have been diagnosed with ADHD type 1,2, or 3. The remaining had their diagnosis withheld for the purpose of a prediction competition. The number of images for each subject ranges from 76 to 276. 264 seed regions of interest are used to define nodes for graphical model analysis (Power et al., 2011). A limited set of covariates including gender, age, handedness, IQ, are available.

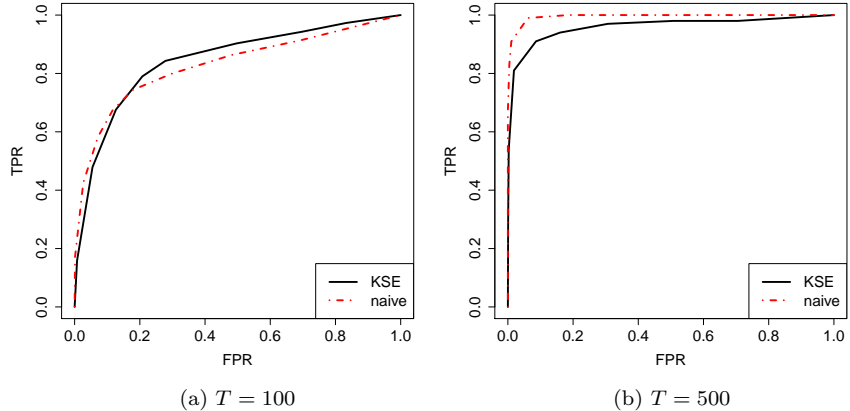

(a) $T = 100$      (b) $T = 500$

Figure 4: ROC curves of KSE and naive under Setting 1: sequentially evolving edges. We set dimension $d = 50$; number of labels $n = 3$; number of pre-fixed edges $n_{\text{fix}=100}$; number of growing edges $n_{\text{grow}} = 500$.

### 4.5.1 Brian Development

In this section, focus lies on investigating the development of brain connectivity network over age for control subjects. Here the subject ages are normalized to be in $[0, 1]$, and the brain ROI measurements are centered to have sample means zero and scaled to have sample standard deviations 1. The bandwidth parameter is set at $h = 0.5$. The regularization parameter $\lambda$ is manually chosen to induce high sparsity for better visualization and highlighting the dominating edges. Consider estimating the brain networks at ages 7.09, 11.75, and 21.83, which are the minimal, median, and maximal ages in the data. Figure 5 shows coronal, sagittal, and transverse snapshots of the estimated brain connectivity networks.

There are two main patterns worth noting in this experiment: (i) It is observed that the degree of complexity of the brain network at the occipital lobe is high compared to other regions by age seven. This is consistent with early maturation of visual and vision processing networks relative to others. We found that this conjecture is supported by several recent scientific results (Shaw et al., 2008; Blakemore, 2012). For example, Shaw et al. (2008) showed that occipital lobe is fully developed before other brain regions. Moreover, when considering structural development, the occipital lobe reaches its peak thickness by age nine. In comparison, portions of the parietal lob reaches their peak thickness as late as thirteen. (ii) Figure 5 also shows that dense connections in the temporal lobe only occur in the graph at age 21.83 among the ages shown. This is also supported by the scientific finding that grey matter in the temporal lobe doesn't reach maximum volume untill age 16 (Bartzokis et al., 2001; Giedd et al., 1999). We also noticed that several confounding factors, such as scanner noise, subject motion, and coregistration, can have potential effects on inference (Braun et al., 2012; Van Dijk et al., 2012). In this manuscript, we rely on the standard data pre-processing techniques as described in Eloyan et al. (2012) for removing such confounders. The influence of these confounders on our inference will be investigated in more details in the future.

Next, we investigate how brain network density changes with age. The number of edges in the estimated graph is controlled by $\lambda$. As Theorem 3.4 indicates, the proper choice of $\lambda$ across the age spectrum depends on the heterogeneity of the multiple time series available. In detail, both the distribution of the subject ages and the number of observations under each subject affect the proper choice of $\lambda$. In order that the same $\lambda$ is applicable across the age spectrum, we take a pre-processing step to achieve homogeneity.

To control the number of observations, $T$, we select the subjects with no fewer than 120 scans. We use only the first 120 scans of these subjects. To make sure that the subjects are distributed uniformly across the age spectrum, we subsampled 46 of the selected subjects whose ages form an equally spaced grid between 10 and 15. We abandon the ranges $[7.09, 10]$ and $[15, 21.83]$, since subjects are distributed rather heterogeneously across these ranges and do not fit into the grid.

Using the subsample of subjects, we can fix $\lambda$ and estimate the brain networks at 26 target ages equally spaced across $[11, 14]$. We do not target at ages close to the boundaries, because fewer subjects are available around these boundaries. Figure 6 demonstrates the estimated number of edges as a function of age, under three choices of $\lambda$. We observe that the estimated brain network density grows with age.

### 4.5.2 The Impact of Bandwidth

In this section, the impact of bandwidths on estimation is considered. In practice, the bandwidth can be regarded as the degree of tradeoff between the label-specific networks and the population level networks. Under such a logic, a higher value of bandwidth will result in incorporating more information from the data points in other labels, and lead to an estimate closer to a population-level graph. This population-level graph will highlight the similarity between different graphs, while tending to ignore the label-specific differences. To illustrate this phenomenon empirically, consider estimating the brain network at age 21.83. We increase the bandwidth $h$, while setting all the other parameters fixed. As $h$ is increased from 0.5 to 3, the weights in Equation (2.4) tends to be homogeneous across ages. Thus the graph ranges from age-specific level to the population level. Figure 7 plots

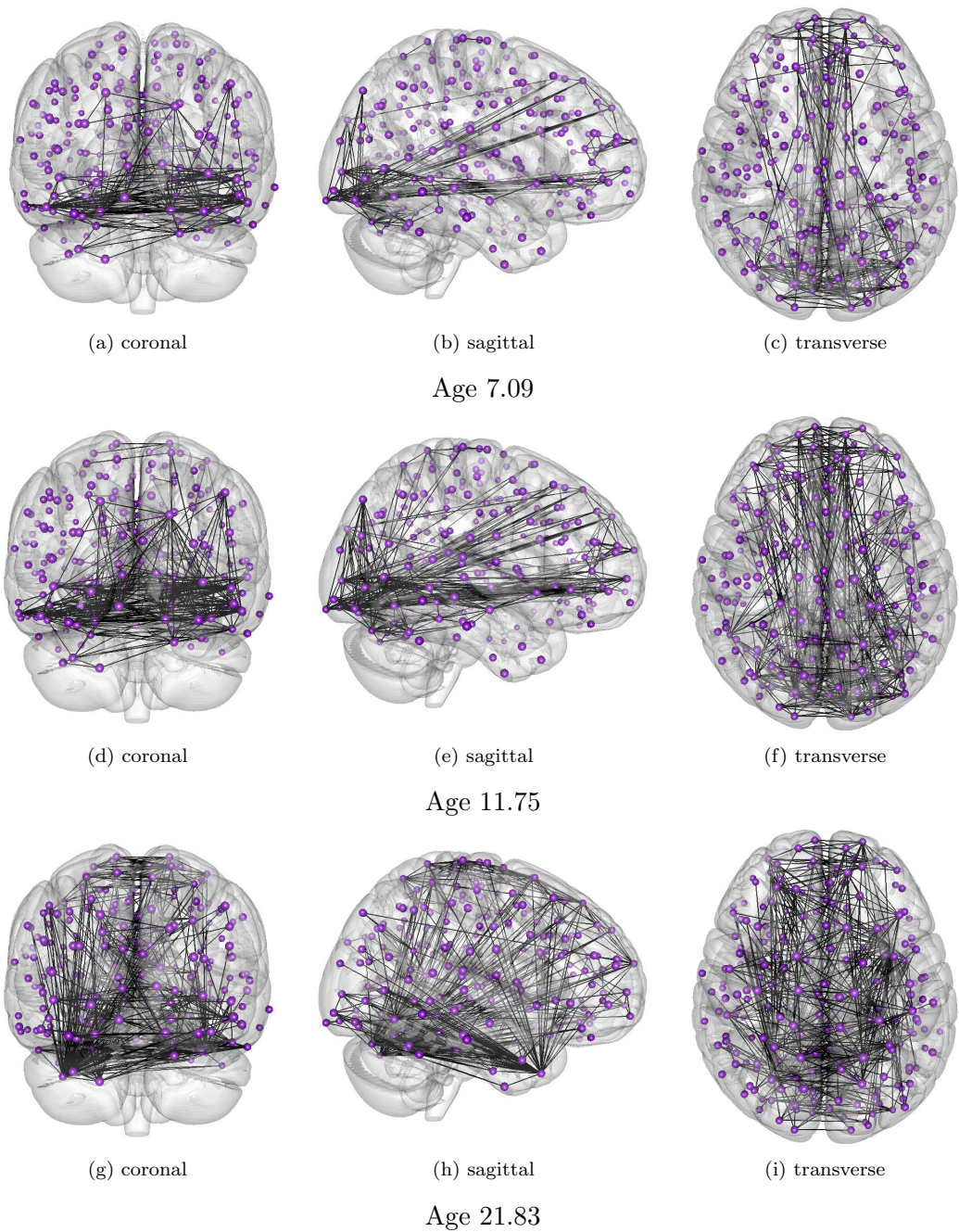

Figure 5: Estimated brain connectivity network at ages 7.09, 11.75, 21.83 among healthy subjects.

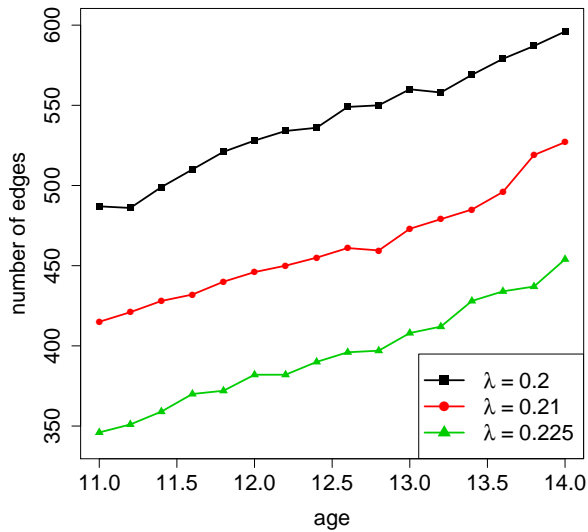

Figure 6: The growth of estimated brain network density over age under three choices of $\lambda$. A subsample of the subjects from the ADHD-200 data are used to control $\lambda$.

the different brain connectivity graphs estimated using different bandwidths.

There are two main discoveries: (i) The number of edges decreases to a population level of 674 as $h$ increase to 3. This is intuitive, because the population level brain network will summarize the information across different levels and thus should be more concrete. (ii) When $h = 3$, the estimated brain network is close to the network estimated at age 7.09 shown in Figure 5 with most edges taking place at the occipital lobe region. This is expected because the occipital lobe region is the only part that has been well developed across the entire range of ages.

# 5 Discussion

In this paper we introduced a new kernel based estimator for jointly estimating multiple graphical models under the condition that the models smoothly vary according to a label. Methodologically, motivated by resting state functional brain connectivity analysis, we proposed a new model, taking both heterogeneity structure and dependence issues into consideration, and introduced a new kernel based method under this model. Theoretically, we provided the model selection and parameter estimation consistency result for the proposed method under both the independence and dependence assumptions. Empirically, we

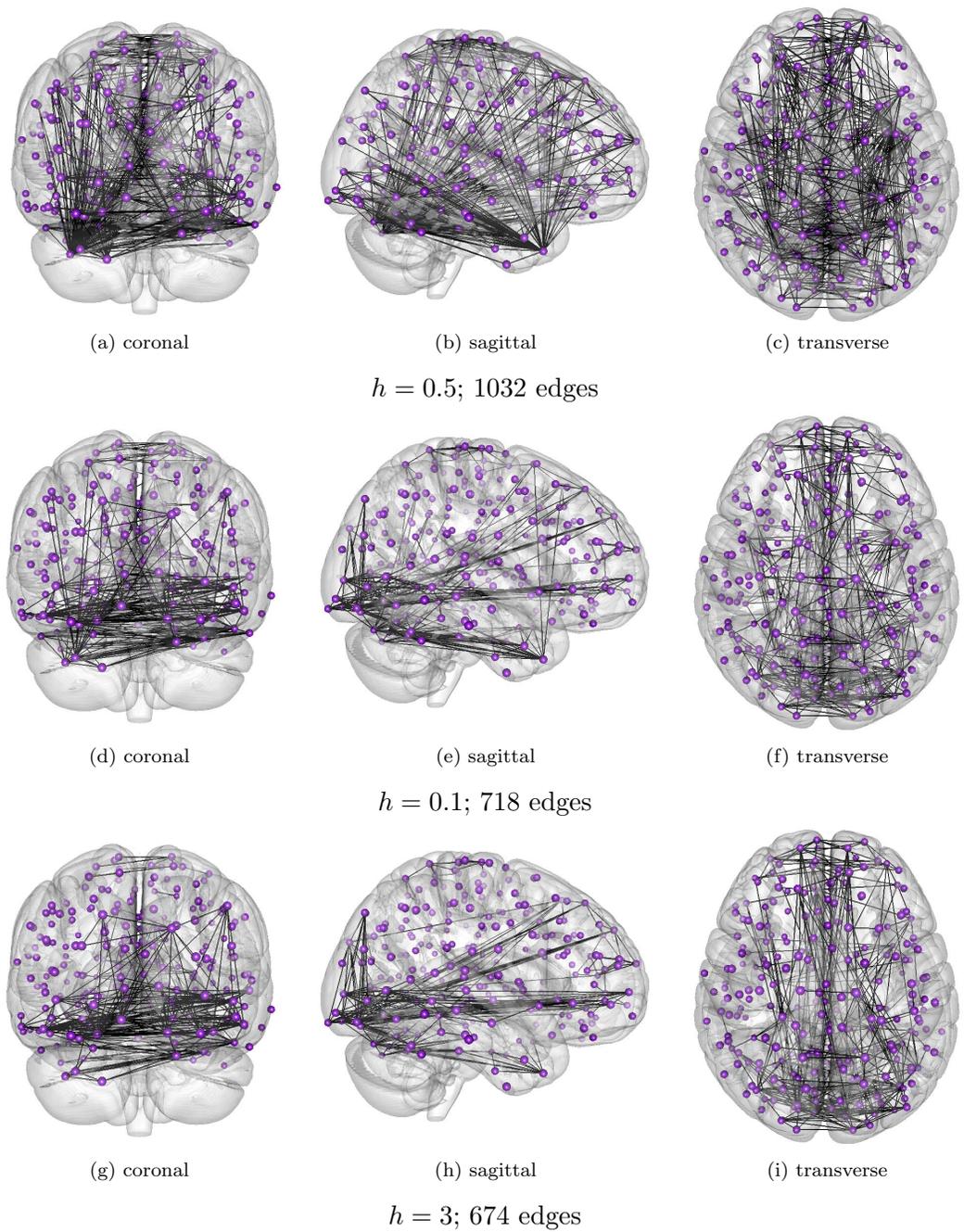

(a) coronal  (b) sagittal  (c) transverse

$h = 0.5$; 1032 edges

(d) coronal  (e) sagittal  (f) transverse

$h = 0.1$; 718 edges

(g) coronal  (h) sagittal  (i) transverse

$h = 3$; 674 edges

Figure 7: Estimated brain connectivity network at age 21.83 among healthy subjects. The kernel bandwidth $h$ takes the value 0.5, 1, 3, resulting to different brain connectivity networks from closer to the age-specific level, to closer to the population level.

applied the proposed method to synthetic and real brain image data. We found that the proposed method is effective for both parameter estimation and model selection compared to several existing methods under various settings.

## Acknowledgement

We would like to thank John Muschelle for providing the R tools to visualize the brain network. We would also like to thank one anonymous referee, the associate editor, and the editor for their helpful comments and suggestions. In addition, thanks also to Drs. Mladen Kolar, Derek Cummings, Martin Lindquist, Michelle Carlson, and Daniel Robinson for helpful discussions on this work.

## A  Proof of Lemma 3.1

The proof of Lemma 3.1 can be decomposed into two parts. In the first part, we prove that the bias term, $\mathbb{E}\mathbf{S}(u_0) - \mathbf{\Sigma}(u_0)$, can be controlled by the number of subjects $n$ and bandwidth $h$. The result is provided in the following lemma.

**Lemma A.1.** *Supposing that the conditions in Lemma 3.1 hold, we have*

$$\max_{j,k} \left| \mathbb{E}\{\mathbf{S}(u_0)\}_{jk} - \Sigma_{jk}(u_0) \right| = O\left(h + \frac{1}{n^2 h^{1+\eta}}\right).$$

*Proof.* By the definition of $\mathbf{S}(u_0)$ in Equation (2.3), we have

$$\mathbf{S}(u_0) = \sum_{i=1}^{n} \omega_i(u_0, h) \frac{1}{T} \sum_{k=1}^{T} \boldsymbol{x}_{ik} \boldsymbol{x}_{ik}^{\top}.$$

Accordingly, we have

$$\begin{aligned}
\mathbb{E}[\mathbf{S}(u_0)]_{jk} &= \sum_{i=1}^{n} \omega_i(u_0, h) \frac{1}{T} \sum_{k=1}^{T} \mathbb{E}\boldsymbol{x}_{ik} \boldsymbol{x}_{ik}^{\top} \\
&= \sum_{i=1}^{n} \omega_i(u_0, h) \Sigma_{jk}(u_i) \\
&= \frac{c(u_0)}{nh} \sum_{i=1}^{n} K\left(\frac{u_i - u_0}{h}\right) \Sigma_{jk}(u_i).
\end{aligned} \quad (A.1)$$

By Theorem 1.1 in Tasaki (2009) and Assumption **(A2)**, we have

$$\frac{c(u_0)}{nh} \sum_{i=1}^{n} K\left(\frac{u_i - u_0}{h}\right) \Sigma_{jk}(u_i)$$

$$= \frac{c(u_0)}{h} \int_0^1 K\left(\frac{u - u_0}{h}\right) \Sigma_{jk}(u) du + O\left[\frac{c(u_0)}{n^2 h} \sup_{u \in [0,1]} \frac{d^2}{du^2}\left\{K\left(\frac{u - u_0}{h}\right) \Sigma_{jk}(u)\right\}\right]$$

$$= c(u_0) \int_{-\frac{u_0}{h}}^{\frac{1-u_0}{h}} K(u) \Sigma_{jk}(u_0 + hu) du + O\left(\frac{1}{n^2 h^{1+\eta}}\right)$$

$$= c(u_0) \int_{a(u_0)}^{b(u_0)} K(u) \left\{\Sigma_{jk}(u_0) + hu \Sigma'_{jk}(\zeta)\right\} du + O\left(\frac{1}{n^2 h^{1+\eta}}\right), \tag{A.2}$$

where $a(u_0) := -I(u_0 \in (0,1])$, $b(u_0) := I(u_0 \in [0,1))$, $\Sigma'_{jk}(u) := \frac{d}{du} \Sigma_{jk}(u)$, and $\zeta$ lies between $u_0$ and $u_0 + hu$. The last equality is because $h \to 0$ and $K(u)$ has support $[-1, 1]$.

By Equation (2.2), we have

$$c(u_0) \int_{a(u_0)}^{b(u_0)} K(u) \Sigma_{jk}(u_0) du = \Sigma_{jk}(u_0). \tag{A.3}$$

By Equation (2.2) and Assumption **(A1)**, we have

$$\left|c(u_0) \int_{a(u_0)}^{b(u_0)} K(u) hu \Sigma'_{jk}(\zeta) du\right| \leq C_2 h \left|c(u_0) \int_{a(u_0)}^{b(u_0)} |u| K(u) du\right|$$

$$= 2 C_2 h \left|\int_0^1 u K(u) du\right| = O(h). \tag{A.4}$$

Combining (A.1), (A.2), (A.3), and (A.4), we have

$$\left|\mathbb{E}\{\mathbf{S}(u_0)\}_{jk} - \Sigma_{jk}(u_0)\right| = O\left(h + \frac{1}{n^2 h^{1+\eta}}\right).$$

This completes the proof. □

We then proceed to the second lemma, which provides an upper bound of the distance between the estimator $\mathbf{S}(u_0)$ and its expectation $\mathbb{E}\mathbf{S}(u_0)$.

**Lemma A.2.** *Supposing that the conditions in Lemma 3.1 hold, we have*

$$\max_{j,k} \left|\{\mathbf{S}(u_0)\}_{jk} - \mathbb{E}\{\mathbf{S}(u_0)\}_{jk}\right| = O_P\left[\frac{\xi \cdot \sup_{u \in [0,1]} \|\mathbf{\Sigma}(u)\|_2}{h\{1 - \sup_{u \in [0,1]} \|\mathbf{A}(u)\|_2\}} \sqrt{\frac{\log d}{Tn}}\right].$$

*Proof.* For $i = 1, \ldots, n$ and $t = 1, \ldots, T$, let $\boldsymbol{y}_{it} := (y_{it1}, \ldots, y_{itd})^\mathsf{T}$ be a $d$-dimensional random vector with $y_{itj} = x_{itj}/\sqrt{\Sigma_{jj}(u_i)}$. Define correlation coefficient $\rho_{jk}(u_i) := \Sigma_{jk}(u_i)/\sqrt{\Sigma_{jj}(u_i)\Sigma_{kk}(u_i)}$. We then have

$$\mathbb{P}\left[|\{\mathbf{S}(u_0)\}_{jk} - \mathbb{E}\{\mathbf{S}(u_0)\}_{jk}| > \epsilon\right]$$

$$= \mathbb{P}\left[\left|\sum_{i=1}^n \omega_i(u_0, h)\left\{\frac{1}{T}\sum_{t=1}^T x_{itj}x_{itk} - \Sigma_{jk}(u_i)\right\}\right| > \epsilon\right]$$

$$= \mathbb{P}\left\{\left|\sum_{i=1}^n \omega_i(u_0, h)\sqrt{\Sigma_{jj}(u_i)\Sigma_{kk}(u_i)}\left(\left[\frac{1}{T}\sum_{t=1}^T (y_{itj} + y_{itk})^2 - 2\{1 + \rho_{jk}(u_i)\}\right]\right.\right.\right.$$

$$\left.\left.\left. - \left[\frac{1}{T}\sum_{t=1}^T (y_{itj} - y_{itk})^2 - 2\{1 - \rho_{jk}(u_i)\}\right]\right)\right| > 4\epsilon\right\}$$

$$\leq \mathbb{P}\left\{\left|\sum_{i=1}^n \omega_i^*(u_0, h)\left(\left[\frac{1}{T}\sum_{t=1}^T (y_{itj} + y_{itk})^2 - 2\{1 + \rho_{jk}(u_i)\}\right]\right)\right| > 2\epsilon\right\}$$

$$+ \mathbb{P}\left\{\left|\sum_{i=1}^n \omega_i^*(u_0, h)\left(\left[\frac{1}{T}\sum_{t=1}^T (y_{itj} - y_{itk})^2 - 2\{1 - \rho_{jk}(u_i)\}\right]\right)\right| > 2\epsilon\right\}$$

$$:= P_1 + P_2, \tag{A.5}$$

where $\omega_i^*(u_0, h) := \omega_i(u_0, h)\sqrt{\Sigma_{jj}(u_i)\Sigma_{kk}(u_i)}$.

Let $\boldsymbol{Z} := (\boldsymbol{Z}_1^\mathsf{T}, \ldots, \boldsymbol{Z}_n^\mathsf{T})^\mathsf{T} \in \mathbb{R}^{nT}$, where $\boldsymbol{Z}_i := (y_{i1j} + y_{i1k}, y_{i2j} + y_{i2k}, \ldots, y_{iTj} + y_{iTk})^\mathsf{T}$. We have $\boldsymbol{Z}_{i_1}$ is independent of $\boldsymbol{Z}_{i_2}$ for any $i_1 \neq i_2$. Let

$$\mathbf{B} := \begin{pmatrix} \sqrt{\omega_1^*(u_0, h)} \cdot \mathbf{I}_T & 0 & \ldots & 0 \\ 0 & \sqrt{\omega_2^*(u_0, h)} \cdot \mathbf{I}_T & & 0 \\ & & \ddots & \\ 0 & 0 & & \sqrt{\omega_n^*(u_0, h)} \cdot \mathbf{I}_T \end{pmatrix}$$

be a $Tn$ by $Tn$ diagonal matrix. Then we can rewrite $P_1$ as $P_1 = \mathbb{P}(|\ \|\mathbf{B}\boldsymbol{Z}\|_2^2 - E\|\mathbf{B}\boldsymbol{Z}\|_2^2\ | > 2T\epsilon)$. Using the property of Gaussian distribution, we have $\mathbf{B}\boldsymbol{Z} \sim N_{Tn}(\mathbf{0}, \mathbf{Q})$, where $\mathbf{Q} := \mathbf{B}\mathrm{cov}(\boldsymbol{Z})\mathbf{B}$ and

$$\mathrm{cov}(\boldsymbol{Z}) = \begin{pmatrix} \mathrm{cov}(\boldsymbol{Z}_1) & 0 & \ldots & 0 \\ 0 & \mathrm{cov}(\boldsymbol{Z}_2) & & 0 \\ & & \ddots & \\ 0 & 0 & & \mathrm{cov}(\boldsymbol{Z}_n) \end{pmatrix}.$$

Let $\{\text{cov}(\boldsymbol{Z}_i)\}_{pq}$ be the $(p,q)$ element of $\text{cov}(\boldsymbol{Z}_i)$. We have

$$\begin{aligned}
|\{\text{cov}(\boldsymbol{Z}_i)\}_{pq}| &= |\text{cov}(y_{ipj} + y_{ipk}, y_{iqj} + y_{iqk})| \\
&= |\text{cov}(y_{ipj}, y_{iqj}) + \text{cov}(y_{ipj}, y_{iqk}) + \text{cov}(y_{ipk}, y_{iqj}) + \text{cov}(y_{ipk}, y_{iqk})| \\
&\leq \frac{|\text{cov}(x_{ipj}, x_{iqj}) + \text{cov}(x_{ipj}, x_{iqk}) + \text{cov}(x_{ipk}, x_{iqj}) + \text{cov}(x_{ipk}, x_{iqk})|}{\min_r \Sigma_{rr}(u_i)} \\
&\leq \frac{4\|\boldsymbol{A}(u_i)\|_2^{|p-q|}\|\boldsymbol{\Sigma}(u_i)\|_2}{\min_r \Sigma_{rr}(u_i)}.
\end{aligned}$$

The last inequality is due to the property of the VAR(1) models. Thus

$$\begin{aligned}
\|\boldsymbol{Q}\|_2 &\leq \max_{1 \leq s \leq Tn} \sum_{r=1}^{Tn} |\boldsymbol{Q}_{sr}| \\
&= \max_{i=1,\ldots,n; p=1,\ldots,T} \sum_{q=1}^{T} \omega_i^*(u_0, h) |\{\text{cov}(\boldsymbol{Z}_i)\}_{pq}| \\
&\leq \max_{i=1,\ldots,n} \omega_i^*(u_0, h) \frac{4\|\boldsymbol{\Sigma}(u_i)\|_2}{\min_r \Sigma_{rr}(u_i)} \cdot 2 \sum_{q=0}^{\infty} \|\boldsymbol{A}(u_i)\|_2^q \\
&\leq \frac{16 C_1}{nh} \cdot \frac{\xi \sup_{u \in [0,1]} \|\boldsymbol{\Sigma}(u)\|_2}{1 - \sup_{u \in [0,1]} \|\boldsymbol{A}(u)\|_2}. \quad (A.6)
\end{aligned}$$

The last inequality is due to the fact that $\omega_i^*(u_0, h) = \omega_i(u_0, h)\sqrt{\Sigma_{jj}(u_i)\Sigma_{kk}(u_i)} \leq \frac{2}{nh} \cdot \sup_v K(v) \cdot \sup_u \max_r \Sigma_{rr}(u)$.

Finally, using Lemma I.2 in Negahban and Wainwright (2011), we have

$$\begin{aligned}
\mathbb{P}(|\ \|\boldsymbol{BZ}\|_2^2 - \mathbb{E}\|\boldsymbol{BZ}\|_2^2\ | > 2T\epsilon) &\leq 2 \exp\left\{-\frac{Tn}{2}\left(\frac{\epsilon}{2n\|\boldsymbol{Q}\|_2} - \frac{2}{\sqrt{Tn}}\right)^2\right\} + 2\exp\left(-\frac{Tn}{2}\right) \\
&\leq 4 \exp\left\{-\frac{Tn}{2}\left(\frac{\epsilon}{4n\|\boldsymbol{Q}\|_2}\right)^2\right\}, \quad (A.7)
\end{aligned}$$

for large enough $n$.

Using the same technique, we can show that $P_2$ in Equation (A.5) can also be controlled by the bound in (A.7). So using the union bound, we have

$$\begin{aligned}
\mathbb{P}\left[\max_{j,k}|\{\boldsymbol{S}(u_0)\}_{jk} - \mathbb{E}\{\boldsymbol{S}(u_0)\}_{jk}| > \epsilon\right] &\leq \sum_{j,k} \mathbb{P}\left[|\{\boldsymbol{S}(u_0)\}_{jk} - \mathbb{E}\{\boldsymbol{S}(u_0)\}_{jk}| > \epsilon\right] \\
&\leq 8d^2 \exp\left(-\frac{T\epsilon^2}{32n\|\boldsymbol{Q}\|_2^2}\right). \quad (A.8)
\end{aligned}$$

Thus, using Equations (A.6) and (A.8), we have

$$\max_{j,k} |\{\mathbf{S}(u_0)\}_{jk} - \mathbb{E}\{\mathbf{S}(u_0)\}_{jk}| = O_P\left(\|\mathbf{Q}\|_2 \sqrt{\frac{n \log d}{T}}\right)$$

$$= O_P\left[\frac{\xi \cdot \sup_{u \in [0,1]} \|\mathbf{\Sigma}(u)\|_2}{h\{1 - \sup_{u \in [0,1]} \|\mathbf{A}(u)\|_2\}} \sqrt{\frac{\log d}{Tn}}\right].$$

This completes the proof. □

*Proof of Lemma 3.1.* The rate of convergence in Lemma 3.1 can be obtained by balancing the convergence rates in Lemmas A.1 and A.2. More specifically, we first have

$$\|\mathbf{S}(u_0) - \mathbf{\Sigma}(u_0)\|_{\max} \leq \|\mathbf{S}(u_0) - \mathbb{E}\mathbf{S}(u_0)\|_{\max} + \|\mathbb{E}\mathbf{S}(u_0) - \mathbf{\Sigma}(u_0)\|_{\max}.$$

For notational brevity, we denote $\theta := \xi \sup_{u \in [0,1]} \|\mathbf{\Sigma}(u)\|_2 / \{1 - \sup_{u \in [0,1]} \|\mathbf{A}(u)\|_2\}$. We then have

$$\|\mathbf{S}(u_0) - \mathbf{\Sigma}(u_0)\|_{\max} = O_P\left(h + \frac{1}{n^2 h^{1+\eta}} + \frac{\theta}{h}\sqrt{\frac{\log d}{Tn}}\right).$$

We first balance the first and third terms in the above upper bound, having that

$$h = \frac{\theta}{h}\sqrt{\frac{\log d}{Tn}} \Rightarrow h = \left(\theta \sqrt{\frac{\log d}{Tn}}\right)^{1/2}.$$

We then balance the first and second terms, and have that

$$h = \frac{1}{n^2 h^{1+\eta}} \Rightarrow h = n^{-\frac{2}{2+\eta}}.$$

Based on the above two results, we have that, on one hand, if $\left(\theta \sqrt{\frac{\log d}{Tn}}\right)^{1/2} > n^{-\frac{2}{2+\eta}}$, we can set

$$h = \left(\theta \sqrt{\frac{\log d}{Tn}}\right)^{1/2}.$$

Then we have

$$h = \frac{\theta}{h}\sqrt{\frac{\log d}{Tn}} > \frac{1}{n^2 h^{1+\eta}} \Rightarrow \|\mathbf{S}(u_0) - \mathbf{\Sigma}(u_0)\|_{\max} = O_P\left\{\left(\theta \sqrt{\frac{\log d}{Tn}}\right)^{1/2}\right\}. \quad (A.9)$$

On the other hand, if $\left(\theta \sqrt{\frac{\log d}{Tn}}\right)^{1/2} \leq n^{-\frac{2}{2+\eta}}$, we can set

$$h = n^{-\frac{2}{2+\eta}}.$$

Then we have

$$h = \frac{1}{n^2 h^{1+\eta}} \geq \frac{\theta}{h}\sqrt{\frac{\log d}{Tn}} \Rightarrow \|\mathbf{S}(u_0) - \mathbf{\Sigma}(u_0)\|_{\max} = O_P\left(n^{-\frac{2}{2+\eta}}\right). \quad (A.10)$$

Combining (A.9) and (A.10), we have the desired result. □

# B  Proof of Theorem 3.2

The following two lemmas are needed in the proof of Theorem 3.2.

**Lemma B.1.** *Let $\mathbf{M}_\rho \in \mathbb{R}^{d \times d}$ be a matrix where $\mathbf{M}_{jk} = \rho^{|j-k|} I(j \neq k)$. Then $\mathbf{M}_\rho$ and $\mathbf{M}_{-\rho}$ have the same set of eigenvalues.*

*Proof.* Let $\mathbf{B} \in \mathbb{R}^{d \times d}$ be a diagonal matrix with $\mathbf{B}_{ii} = (-1)^i$. Noting that $(-1)^{i+j} = (-1)^{|i-j|}$ for all $i, j \in \{1, \ldots, d\}$, we have $\mathbf{M}_{-\rho} = \mathbf{B}\mathbf{M}\mathbf{B}^{-1}$. Thus $\mathbf{M}_\rho$ has the same set of eigenvalues as $\mathbf{M}_{-\rho}$. □

**Lemma B.2.** *Let $\mathbf{N}_\rho \in \mathbb{R}^{d \times d}$ be a matrix where $\mathbf{N}_{jk} = \rho^{|j-k|}$ and $0 \leq \rho_1 \leq \rho_2$, we have $\|\mathbf{N}_{\rho_1}\|_2 \leq \|\mathbf{N}_{\rho_2}\|_2$.*

*Proof.* $\mathbf{N}_{\rho_1}$ is the Hadamard product of $\mathbf{N}_{\rho_1/\rho_2}$ and $\mathbf{N}_{\rho_2}$:

$$\mathbf{N}_{\rho_1} = \mathbf{N}_{\rho_1/\rho_2} \circ \mathbf{N}_{\rho_2}.$$

By Theorem 5.3.4 of Roger and Charles (1994), any eigenvalue $\lambda(\mathbf{N}_{\rho_1/\rho_2} \circ \mathbf{N}_{\rho_2})$ of $\mathbf{N}_{\rho_1/\rho_2} \circ \mathbf{N}_{\rho_2}$ satisfies

$$\lambda(\mathbf{N}_{\rho_1/\rho_2} \circ \mathbf{N}_{\rho_2}) \leq (\max_{1 \leq i \leq d} \mathbf{N}_{\rho_1/\rho_2})_{ii} \lambda_{\max}(\mathbf{N}_{\rho_2}) = \|\mathbf{N}_{\rho_2}\|_2.$$

Thus $\|\mathbf{N}_{\rho_1}\|_2 \leq \|\mathbf{N}_{\rho_2}\|_2$. □

*Proof of Theorem 3.2.* Under Scenario (1), it is straightforward to have $\|\mathbf{A}\|_2 = \max_{j=1,\ldots,d} |\rho_j|$. Plugging it into Equation 3.2 proves the first part.

Under Scenario (2).i, it is well known that $\|\mathbf{A}\|_2 = 2|\rho|\cos\{\pi/(d+1)\}$. See, for example, Smith (1978) for details. This proves the second part.

Under Scenario (2).ii, the eigenvalues of $\mathbf{A}$ consist of the eigenvalues of each block. From Lemma B.1, we conclude that $\|\mathbf{A}\|_2$ do not depend on the sign of $\rho$. To prove monotonicity, note that $\|\mathbf{A}\|_2 = \max_{l=1,\ldots,k} \|\mathbf{A}_l\|_2$ and $\|\mathbf{A}_l\|_2 = \|\mathbf{N}_\rho - I_{d_l}\|_2 = \|\mathbf{N}_\rho\|_2 - 1$ for $\mathbf{N}_\rho \in \mathbb{R}^{d_l \times d_l}$. The desired result follows from Lemma B.2. □

# C  Proof of Lemma 3.3

To prove Lemma 3.3, we need an improved upper bound on the distance between $\mathbf{S}(u_0)$ and $\mathbb{E}\mathbf{S}(u_0)$. We provide such a result in Lemma C.1. The proof of Lemma C.1 can be regarded as an extension to the proof of Lemma 6 in Zhou et al. (2010).

**Lemma C.1.** *Suppose that Assumptions* **(B1)**, **(B2)**, *and* **(B3)** *in Lemma 3.3 hold, and $n^{-2/5} < h < 1$. Then we have there exist absolute positive constants $C_4$ and $C_5$, such that for*

$$\epsilon < \frac{C_4\{\Sigma_{jj}^2(u_0)\Sigma_{kk}^2(u_0) + \Sigma_{jk}^2(u_0)\}}{\max_{i=1,\ldots,n} K\{(u_i - u_0)/h\}\Sigma_{jj}(u_i)\Sigma_{kk}(u_i)},$$

*we have*

$$\mathbb{P}\left[|\{\mathbf{S}(u_0)\}_{jk} - \mathbb{E}\{\mathbf{S}(u_0)\}_{jk}| > \epsilon\right] \leq 2\exp(-C_5 Tnh\epsilon^2).$$

*Proof.* By the definition of $\mathbf{S}(u_0)$, we have

$$\mathbb{P}\left[|\{\mathbf{S}(u_0)\}_{jk} - \mathbb{E}\{\mathbf{S}(u_0)\}_{jk}| > \epsilon\right] = \mathbb{P}\left[\left|\sum_{i=1}^n w_i(u_0, h)\left\{\frac{1}{T}\sum_{t=1}^T x_{itj}x_{itk} - \Sigma_{jk}(u_i)\right\}\right| > \epsilon\right]$$

$$\leq \mathbb{P}\left[\sum_{i=1}^n w_i(u_0, h)\left\{\frac{1}{T}\sum_{t=1}^T x_{itj}x_{itk} - \Sigma_{jk}(u_i)\right\} > \epsilon\right]$$

$$+ \mathbb{P}\left[\sum_{i=1}^n w_i(u_0, h)\left\{-\frac{1}{T}\sum_{t=1}^T x_{itj}x_{itk} + \Sigma_{jk}(u_i)\right\} > \epsilon\right]$$

$$:= P_3 + P_4.$$

By Markov's inequality, $\forall r > 0$,

$$P_3 = \mathbb{P}\left(\exp\left[Tnr\sum_{i=1}^n w_i(u_0, h)\left\{\frac{1}{T}\sum_{t=1}^T x_{itj}x_{itk} - \Sigma_{jk}(u_i)\right\}\right] > e^{Tnr\epsilon}\right)$$

$$\leq \frac{1}{e^{Tnr\epsilon}}\mathbb{E}\exp\left[r\sum_{i=1}^n \frac{2}{h}K\left(\frac{u_i - u_0}{h}\right)\sum_{t=1}^T\{x_{itj}x_{itk} - \Sigma_{jk}(u_i)\}\right]$$

$$= e^{-Tnr\epsilon}\prod_{i=1}^n \exp\left\{-Tr\frac{2}{h}K\left(\frac{u_i - u_0}{h}\right)\Sigma_{jk}(u_i)\right\}\prod_{i=1}^n\left[\mathbb{E}\exp\left\{r\frac{2}{h}K\left(\frac{u_i-u_0}{h}\right)x_{itj}x_{itk}\right\}\right]^T.$$

The last equality is due to that $\{\boldsymbol{X}^{u_i}\}_{i=1}^n$ are independent and $\{\boldsymbol{x}_{it}\}_{t=1}^T$ are i.i.d.. Using the same technique, we can get similar result for $P_4$. The rest of the proof can be derived by following Lemma 6 in Zhou et al. (2010), where we replace $n$ with $Tn$. Here the assumption that $n^{-2/5} < h < 1$ and Assumption **(B2)** are required in the proof of Proposition 5 in Zhou et al. (2010).  □

Using Lemma C.1, we can now proceed to prove Lemma 3.3. Because if the kernel function satisfies Assumption **(A2)** for some $\eta = \eta_1 > 0$, then this kernel function also satisfies Assumption **(A2)** for $\eta = \max(3, \eta_1)$, so without loss of generality, in the sequel we assume that $\eta \geq 3$ in Assumption **(A2)**.

*Proof of Lemma 3.3.* Using Lemma C.1, we have

$$\mathbb{P}\left[\max_{jk} |\{\mathbf{S}(u_0)\}_{jk} - \mathbb{E}\{\mathbf{S}(u_0)\}_{jk}| > \epsilon\right] \leq \sum_{jk} \mathbb{P}\left[|\{\mathbf{S}(u_0)\}_{jk} - \mathbb{E}\{\mathbf{S}(u_0)\}_{jk}| > \epsilon\right]$$

$$\leq \exp\left(2\log d - C_5 T n h \epsilon^2\right),$$

for $n^{-2/5} < h < 1$. Now setting $\epsilon = \sqrt{3\log d/(C_5 T n h)}$, we have

$$\mathbb{P}\left[\max_{jk} |\{\mathbf{S}(u_0)\}_{jk} - \mathbb{E}\{\mathbf{S}(u_0)\}_{jk}| > \sqrt{\frac{3\log d}{C_5 T n h}}\right] \leq \frac{1}{d}.$$

Accordingly, as $d \to \infty$, we have

$$\max_{jk} |\{\mathbf{S}(u_0)\}_{jk} - \mathbb{E}\{\mathbf{S}(u_0)\}_{jk}| = O_P\left(\sqrt{\frac{\log d}{T n h}}\right).$$

Together with Lemma A.1, we have

$$\|\mathbf{S}(u_0) - \mathbf{\Sigma}(u_0)\|_{\max} = O_P\left(h + \frac{1}{n^2 h^{1+\eta}} + \sqrt{\frac{\log d}{T n h}}\right).$$

Similarly as the proof of Lemma 3.1, to balance the first and third terms, we set

$$h = \sqrt{\frac{\log d}{T n h}} \Rightarrow h = \left(\frac{\log d}{T n}\right)^{1/3}.$$

To balance the first and second terms, we set

$$h = \frac{1}{n^2 h^{1+\eta}} \Rightarrow h = \frac{1}{n^{2/(2+\eta)}}.$$

If $\left(\frac{\log d}{T n}\right)^{1/3} > \frac{1}{n^{2/(2+\eta)}}$, we set

$$h = \left(\frac{\log d}{T n}\right)^{1/3}.$$

Then we have

$$h = \sqrt{\frac{\log d}{T n h}} > \frac{1}{n^2 h^{1+\eta}} \Rightarrow \|\mathbf{S}(u_0) - \mathbf{\Sigma}(u_0)\|_{\max} = O_P\left\{\left(\frac{\log d}{T n}\right)^{1/3}\right\}. \quad \text{(C.1)}$$

Note that $\eta \geq 3$ implies that $h > n^{-2/(2+\eta)} > n^{-2/5}$.

If $\left(\frac{\log d}{Tn}\right)^{1/3} \leq \frac{1}{n^{2/(2+\eta)}}$, we set
$$h = \frac{1}{n^{2/(2+\eta)}}.$$

Then we have
$$h = \frac{1}{n^2 h^{1+\eta}} \geq \sqrt{\frac{\log d}{Tnh}} \Rightarrow \|\mathbf{S}(u_0) - \mathbf{\Sigma}(u_0)\|_{\max} = O_P\left\{\frac{1}{n^{2/(2+\eta)}}\right\}. \tag{C.2}$$

Combining (C.1) and (C.2) we have the desired result. $\square$